%% file: main.tex
\pdfoutput=1
\documentclass{zgroup/zgroup}

\usepackage[utf8]{inputenc} 
\usepackage[T1]{fontenc}    
\usepackage{hyperref}       
\usepackage{url}            
\usepackage{booktabs}       
\usepackage{amsfonts}       
\usepackage{nicefrac}       
\usepackage{microtype}      
\usepackage{xcolor}         
\usepackage{bbm}
\usepackage{algorithm}
\usepackage{algorithmic}
\usepackage{graphicx}
\usepackage{xcolor}
\usepackage{colortbl}
\usepackage{natbib}
\usepackage{enumitem}
\usepackage{setspace}
\usepackage{caption}
\usepackage{multirow}
\usepackage{wrapfig}
\usepackage{gensymb}

\newif\ifcomments
\newif\ifarxiv
\arxivtrue
\commentsfalse

\title{
Online Unsupervised Learning of Visual \\Representations and Categories
}

\author{%
Mengye Ren$^{1,2}$ \quad Tyler R. Scott$^{3,4}$ \quad Michael L. Iuzzolino$^{4}$\\
\textbf{Michael C. Mozer}$^{3,4}$ \quad \textbf{Richard Zemel}$^{1,2}$\\
$^{1}$University of Toronto \quad $^{2}$Vector Institute \quad $^{3}$Google \quad $^{4}$University of Colorado, Boulder\\
  \texttt{\{mren,zemel\}@cs.toronto.edu}\\
  \texttt{\{tysc7237,michael.iuzzolino,mozer\}@colorado.edu} \\
}

\zgroupauthor{\Name{Mengye Ren} \Email{mengye@nyu.edu}\\
  \addr New York University; Google \\
  \Name{Tyler R. Scott} \Email{tylersco@google.com}\\
  \addr Google; University of Colorado, Boulder\\
  \Name{Michael L. Iuzzolino} \Email{michael.iuzzolino@colorado.edu}\\
  \addr University of Colorado, Boulder \\
  \Name{Michael C. Mozer} \Email{mcmozer@google.com}\\
  \addr Google; University of Colorado, Boulder\\
  \Name{Richard Zemel} \Email{zemel@cs.columbia.edu}\\
  \addr Columbia University; University of Toronto; Vector Institute; Canadian Institute for Advanced Research\\
}

\input{sections/macro}
\input{sections/math}

\begin{document}

\maketitle
\morespace{-0.3in}
\begin{abstract}
\input{sections/abstract}
\end{abstract}

\input{sections/intro}
\input{sections/related}
\input{sections/model}
\input{sections/experiments}
\input{sections/conclusion}
\input{sections/ack}

{
\bibliography{ref}
}


\clearpage
\appendix
\input{sections/appendix}

\end{document}

%% file: sections/macro.tex
\newcommand{\rom}{{RoamingOmniglot}}
\newcommand{\rim}{{RoamingImageNet}}
\newcommand{\rro}{{RoamingRooms}}

\ifarxiv
\newcommand{\savespace}[1]{}
\newcommand{\savespacefigtop}[1]{}
\newcommand{\morespace}[1]{\vspace{#1}}
\fi
\definecolor{red}{RGB}{255,73,92}
\definecolor{blue}{RGB}{37,110,255}
\definecolor{violet}{RGB}{70,35,122}
\definecolor{green}{RGB}{61,220,151}

%% file: sections/math.tex
\usepackage{amsmath,amsfonts,bm}

\def\eqref#1{equation~\ref{#1}}

\def\1{\bm{1}}

\def\rvp{{\mathbf{p}}}

\def\rvx{{\mathbf{x}}}

\def\rvz{{\mathbf{z}}}

\DeclareMathAlphabet{\mathsfit}{\encodingdefault}{\sfdefault}{m}{sl}
\SetMathAlphabet{\mathsfit}{bold}{\encodingdefault}{\sfdefault}{bx}{n}

\def\gL{{\mathcal{L}}}

\newcommand{\softmax}{\mathrm{softmax}}
\newcommand{\sigmoid}{\mathrm{sigmoid}}
\newcommand{\logsumexp}{\mathrm{logsumexp}}

\DeclareMathOperator*{\argmin}{arg\,min}

%% file: sections/abstract.tex
\looseness=-100
Real world learning scenarios involve a nonstationary distribution of classes with sequential
dependencies among the samples, in contrast to the standard machine learning formulation of drawing
samples independently from a fixed, typically uniform distribution. Furthermore, real world
interactions demand learning on-the-fly from few or no class labels.  In this work, we propose an
unsupervised model that simultaneously performs online visual representation learning and few-shot
learning of new categories without relying on any class labels. Our model is a prototype-based
memory network with a control component that determines when to form a new class prototype. We
formulate it as an online 
mixture model, where components are created with only a
single new example, and assignments do not have to be balanced, which permits an approximation to
natural imbalanced distributions from uncurated raw data. Learning includes a contrastive loss that
encourages different views of the same image to be assigned to the same prototype. The result is a
mechanism that forms categorical representations of objects in nonstationary environments.
Experiments show that our method can learn from an online stream of visual input data and 
its learned representations are significantly better
at category recognition compared to state-of-the-art self-supervised learning
methods.

%% file: sections/intro.tex
\section{Introduction}
\savespace{-0.1in}

\looseness=-100
Humans operating in the real world have the opportunity to learn from large
quantities of unlabeled data. However, as an individual moves within and
between environments, the stream of experience has complex temporal
dependencies. The goal of our research is to tackle the challenging problem of
online unsupervised representation learning in the setting of environments with
naturalistic structure. We wish to design learning algorithms that facilitate
the categorization of objects as they are encountered and re-encountered. In
representation learning, methods are often evaluated based on their ability to
classify from the representation using either supervised linear readout or
unsupervised clustering over the full dataset, both of which are typically done
in a separate post-hoc evaluation phase. Instead, a key aim of our work is to
predict object categories throughout training and evaluation, where
categorization is performed by grouping a new instance with one or more
previous instances, and does not rely on externally provided labels at any
stage.

Unsurprisingly, the structure of natural environments contrasts dramatically
with the standard scenario typically assumed by many machine learning
algorithms: mini-batches of independent and identically distributed (iid)
samples from a well-curated dataset. In unsupervised visual representation
learning, the most successful methods rely on iid samples. Contrastive-based
objectives~\citep{simclr,moco} typically assume that each instance in the
mini-batch forms its own instance class. When this assumption is violated due
to autocorrelations in a naturalistic online streaming setting, contrastive
approaches will push same-class instances apart. Clustering-based learning
frameworks~\citep{deepcluster,sela,swav} have their own difficulties in
environments with nonstationary and imbalanced class distributions: they assume
that the set of cluster centroids remain relatively stable and that the
clusters are balanced in size.

\looseness=-1
To make progress on the challenge of unsupervised visual representation
learning and categorization in a naturalistic setting, we propose the
\emph{online unsupervised prototypical network (OUPN)}, which performs learning
of visual representations and object categories simultaneously in a
single-stage process. Class prototypes are created via an online clustering
procedure, and a contrastive loss~\citep{chopra2005,cpc} is used to encourage
different views of the same image to be assigned to the same cluster. Notably,
our online clustering procedure is more flexible relative to other
clustering-based representation learning algorithms, such as
DeepCluster~\citep{deepcluster} and SwAV~\citep{swav}: OUPN performs learning
and inference as an online Gaussian mixture model, where clusters can be
created online with only a single new example, and cluster assignments do not
have to be balanced, which permits an approximation to natural imbalanced
distributions from uncurated raw data.

\hyphenation{Roam-ing-Rooms}
\looseness=-10000
We train and evaluate our algorithm on a recently proposed naturalistic
dataset, RoamingRooms~\citep{www}, which uses imagery collected from a virtual
agent walking through different rooms, and SAYCam~\citep{saycam}, which is
collected from head-mounted camera recordings from human babies. We compare to
a suite of state-of-the-art self-supervised representation learning methods:
SimCLR~\citep{simclr}, SwAV~\citep{swav}, and SimSiam~\citep{simsiam}. OUPN
performs relatively well, as these methods are designed for batches of iid data
and degrade significantly with non-iid streams.  But even when we train these
methods in an offline fashion---by shuffling the data to be iid---they
underperform OUPN, which handles better the underlying data imbalance and
exploits structure in the online temporal streams. In addition, we use
RoamingOmniglot~\citep{www} as a benchmark, and also investigate the effect of
imbalanced classes; we find that OUPN is very robust to an imbalanced
distribution of classes. For a version of ImageNet with non-iid structure,
RoamingImageNet, OUPN again outperforms self-supervised learning baselines when
using matched batch sizes. These experiments indicate that OUPN supports the
emergence of visual understanding and category formation of an online agent
operating in an embodied environment.

%% file: sections/related.tex
\section{Related Work}
\savespace{-0.1in}
\paragraph{Self-supervised learning.} 
\looseness=-10000 Self-supervised learning methods discover
rich and informative visual representations without class labels.
\textit{Instance-based approaches} aim to learn invariant representations of
each image under different
transformations~\citep{cpc,pirl,cmc,moco,simclr,mocov2,byol,simsiam}. They
typically work well with iid data and large batch sizes, which contrasts with
realistic learning scenarios. Our method is also related to
\textit{clustering-based approaches}, which obtain clusters on top of the
learned embedding and use the cluster assignments to constrain the embedding
network. To compute the cluster assignment,
DeepCluster~\citep{deepcluster,onlinedeepcluster} and PCL~\citep{pcl} use the
$k$-means algorithm whereas SeLa~\citep{sela} and SwAV~\citep{swav} uses the
Sinkhorn-Knopp algorithm~\citep{sinkhornknopp}. However, they typically assume
a fixed number of clusters, and Sinkhorn-Knopp further assumes a balanced
assignment as an explicit constraint. In contrast, our online clustering
procedure is more flexible: it can create new clusters on-the-fly with only a
single new example and does not assume balanced cluster assignments.
Self-supervised pretraining or joint training has proven beneficial for online
continual learning tasks \citep{zhang2020self,gallardo2021self,cha2021co2l}.

\savespace{-0.12in}
\paragraph{Representation learning from video.} 
There has also been a surge of interest in leveraging video data to learn
visual
representations~\citep{unsupvideo,pathak2017learning,eyesofachild,s3vae,flowe}.
These approaches all sample video subsequences uniformly over the entire
dataset, whereas our model directly learns from an online stream of data. Our
model also does not have the assumption that inputs must be adjacent frames in
the video.

\savespace{-0.12in}
\paragraph{Online and incremental representation learning.} Our work is also
related to online and continual representation
learning~\citep{icarl,eeil,curl,onlinemixturetasks,oml,remind}. Continual
mixture models~\citep{curl,onlinemixturetasks} designate a categorical latent
variable that can be dynamically allocated for a new environment. Our model has
a similar mixture latent variable setup but one major difference is that we
operate on example-level rather than task-level. Streaming
learning~\citep{stream,remind} aims to perform representation learning online.
Most work here except~\citet{curl} assumes a fully supervised setting. Our
prototype memory also resembles a replay buffer~\citep{darker,prs}, but we
store the feature prototypes instead of the inputs.

\savespace{-0.12in}
\paragraph{Latent variable modeling on sequential data.} Our model also relates
to a family of latent variable generative models for sequential
data~\citep{dkf,composinggraphical,videovae,videogenvae,s3vae}. Like our model,
these approaches aim to infer latent variables with temporal structure, but
they use an input reconstruction criterion.

\savespace{-0.12in}
\paragraph{Online mixture models.} Our clustering module is related to the
literature on online mixture models,
e.g.,~\cite{art,humancategory,bottou1995convergence,song2005highly,memoizeddp,fastgmm}.
Typically, these are designed for fast and incremental learning of clusters
without having to recompute clustering over the entire dataset. Despite
presenting a similar online clustering algorithm, our goal is to jointly learn
both online clusters and input representations that facilitate future online
clustering episodes.

\savespace{-0.1in}
\paragraph{Few-shot learning.} 
\looseness=-10000
Our model can recognize new classes with only one or a few examples. Our
prototype-based memory is also inspired by the Prototypical Network and its
variants \citep{protonet,imp,www}. Few-shot methods can reduce or remove
reliance on class labels using semi- and self-supervised
learning~\citep{fewshotssl,centroidnet,cactus,boostingfewshot,aal,umtra,protoclr}.

\looseness=-10000
Classical few-shot learning, however, relies on episodes of equal number of
training and test examples from a fixed number of new classes.
\citet{dynamicfewshot,metadataset,tao2020few,zhu2021prototype} consider
extending the standard episodes with incremental learning and varying number of
examples and classes. \citet{www} proposed a new setup that incrementally
accumulates new classes and re-visits old classes over a sequence of inputs. We
evaluate our algorithm on a similar setup; however, unlike that work, our
proposed algorithm does not rely on any class labels.

\savespace{-0.1in}
\paragraph{Human category learning.} 
\looseness=-10000
Our work is related to human learning settings and online clustering models
from cognitive
science~\citep{art,conceptformation,humancategory,sustain,bigbook,lakecategory}.
These models assume a known, fixed representation of inputs. In contrast, our
model learns both representations and categories in an end-to-end fashion.

%% file: sections/model.tex
\begin{figure}
\savespacefigtop{-0.5in}
\centering
\ifarxiv
\includegraphics[width=\textwidth,trim={-0.5cm 9cm 2.2cm 0cm},clip]{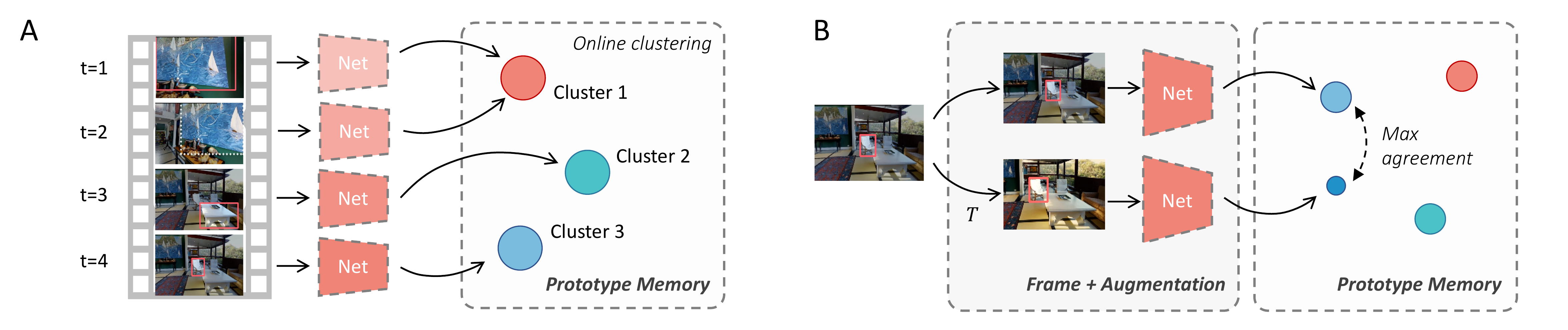}
\fi
\savespace{-0.1in}
\caption{Our proposed online unsupervised prototypical network (OUPN).
\textbf{A:} OUPN learns directly from an online visual stream. Images are
processed by a deep neural network to extract representations. Representations
are stored and clustered in a prototype memory. Similar features are aggregated
in a cluster and new clusters can be dynamically created if the current feature
vector is different from all existing clusters. \textbf{B:} The network
learning uses self-supervision that encourages different augmentations of the
same frame to have consistent cluster assignments.}
\savespace{-0.2in}
\end{figure}

\section{Online Unsupervised Prototypical Networks}
\savespace{-0.1in}
We now introduce our model, \emph{online unsupervised prototypical networks}
(\emph{OUPN}), which operates in a streaming categorization setting. At each
time step $t$, OUPN receives an input $\rvx_t$ and predicts both a categorical
variable $\hat{y}_t$ that indicates the object class and also a binary variable
$\hat{u}_t$ that indicates whether the class is known ($u=0$) or new ($u=1$).
OUPN  uses a network $h$ to encode the input to obtain embedding $\rvz_t =
h(\rvx_t; \theta)$, where $\theta$ represents the learnable parameters of the
encoder network.

\looseness=-10000
We first describe the inference procedure to cluster embeddings obtained by a
fixed $\theta$ using an online probabilistic mixture model. Next, we propose a
multi-component loss for representation learning in our setting which allows
$\theta$ to be learned from scratch in the course of online clustering.

\savespace{-0.1in}
\subsection{Inference}
\label{sec:inference}
\savespace{-0.1in}
We formulate our clustering inference procedure in terms of a probabilistic
mixture model, where each cluster corresponds to a Gaussian distribution
$f(\rvz; \rvp, \sigma^2)$, with mean $\rvp$, a constant isotropic variance
$\sigma^2$ shared across all clusters, and mixture weights $w$: $p(\rvz; P) =
\sum_k w_k f(\rvz; \rvp_k, \sigma^2).$ Throughout a sequence, the number of
components evolves as the model makes an online decision of when to create a
new cluster or remove an old one. We assume that the prior distribution for the
Bernoulli variable $u$ is constant---$u_0 \equiv \Pr(u=1) $)---and the prior
for a new cluster is uniform over the entire space---$z_0 \equiv \Pr(\rvz |
u=1) $ \citep{lathuiliere2018deepgum}. In the following, we characterize
inference as an approximate extension of the EM algorithm to a streaming
setting. The full derivation is included in Appendix~\ref{sec:derivation}.

\savespace{-0.1in}
\subsubsection{E-step}
\savespace{-0.1in}
Upon seeing the current input $\rvz_t$, the online clustering procedure needs
to predict the cluster assignment or initiate a new cluster in the E-step.

\savespace{-0.12in}
\paragraph{Inferring cluster assignments.}
The categorical variable $\hat{y}$ infers the cluster assignment of the current
input example with regard to the existing clusters.
\ifarxiv
\begin{align}
    \hat{y}_{t,k} &= \Pr(y_t = k | \rvz_{t}, u=0) = \frac{\Pr(\rvz_{t} | y_t=k, u=0) \Pr(y_t=k)}{\Pr(\rvz_{t}, u=0)} \\
    &= \frac{w_k f(\rvz_{t};
    \rvp_{t,k}, \sigma^2)}{\sum_{k'} w_{k'} f(\rvz_{t}; \rvp_{t,k'}, \sigma^2)} 
    = \softmax
    \left(\log w_k - \frac{1}{\tau} d(\rvz_t, \rvp_{t,k}) \right),
\end{align}
\fi
where $w_k$ is the mixing coefficient of cluster $k$, $d(\cdot,\cdot)$ is the
distance function, and $\tau$ is an independent learnable temperature parameter
that is related to the cluster variance.

\savespace{-0.12in}
\paragraph{Inference on unknown classes.} The binary variable $\hat{u}$
estimates the probability that the current input belongs to a new cluster:
\ifarxiv
\begin{align}
    \hat{u}_t &= \Pr(u_t = 1 | \rvz_{t})\\
    &= \frac{z_0 u_0}{z_0 u_0 + \sum_k w_k f(\rvz_t; \rvp_{t,k}, \sigma^2) (1-u_0)}\\
    &\ge \frac{z_0 u_0}{z_0 u_0 + \max_k f(\rvz_t; \rvp_{t,k}, \sigma^2) (1-u_0)} \\
    \label{eq:maxapprox}
    &= \sigmoid((\min_k \frac{1}{\tau} d(\rvz_t, \rvp_{t,k}) - \beta) / \gamma),
\end{align}
\fi
where $\beta$ and $\gamma$ are separate learnable parameters related to $z_0$
and $u_0$, allowing us to predict different confidence levels for unknown and
known classes.

\savespace{-0.1in}
\subsubsection{M-step}
\savespace{-0.1in}
\label{sec:mstep}
Here we infer the posterior distribution of the cluster centroids $\Pr(\rvp_{t,k} |
\rvz_{1:t})$. We formulate an efficient recursive online update, similar to
Kalman filtering, incorporating the evidence of the current input $\rvz_t$ and
avoiding re-clustering the entire input history. We define $\hat{\rvp}_{t,k}$
as the posterior estimate of the mean of the $k$-th cluster at time step $t$,
and $\hat{c}_{t,k}$ is the estimate of the inverse variance.

\savespace{-0.1in}
\paragraph{Updating centroids.}
Suppose that in the E-step we have determined that $y_t=k$. Then the posterior
distribution of the $k$-th cluster after observing $\rvz_t$ is:
\begin{align*}
    &\Pr(\rvp_{t,k} | \rvz_{1:t}, y_t=k)
    \propto \Pr( \rvz_{t} | \rvp_{t,k}, y_t=k) \Pr(\rvp_{t,k} | \rvz_{1:t-1})\\
    &\approx f(\rvz_t; \rvp_{t,k}, \sigma^2) \int_{\rvp'} 
    f(\rvp_{t,k}; \rvp', \sigma_{t,d}^2) f(\rvp'; \hat{\rvp}_{t-1,k}, \hat{\sigma}_{t-1,k}^2) \\
    &= f(\rvz_t; \rvp_{t,k}, \sigma^2)
    f(\rvp_{t,k}; \hat{\rvp}_{t-1,k}, \sigma_{t,d}^2 + \hat{\sigma}_{t-1,k}^2).
\end{align*}
The transition probability distribution $\Pr(\rvp_{t,k} | \rvp_{t-1,k})$ is a
zero-mean Gaussian with variance $\hat{\sigma}_{t,d}^2 = (1/\rho - 1)
\hat{\sigma}_{t-1,k}^2$, where $\rho \in (0, 1]$ is some constant that we
define to be the memory decay coefficient. Since the representations are learnable, we assume that $\sigma^2=1$, and
the memory update equation can be formulated as
follows:
\ifarxiv
\begin{align}
\hat{c}_{t,k} &= \mathop{\mathbb{E}}_{y_t} [\hat{c}_{t,k}|_{y_t}]
= \rho \hat{c}_{t-1,k} + \hat{y}_{t,k} (1-\hat{u}_{t,k}); \\
\hat{\rvp}_{t,k} &= \mathop{\mathbb{E}}_{y_t} [\hat{\rvp}_{t,k} |_{y_t}] 
= \rvz_t \frac{\hat{y}_{t,k} (1-\hat{u}_{t,k})}{\rho\hat{c}_{t-1,k} + 1} + 
\hat{\rvp}_{t-1,k} \left(1-  \frac{\hat{y}_{t,k}(1-\hat{u}_{t,k})}{\rho\hat{c}_{t-1,k} + 1} \right);\\
\hat{w}_{t,k} &= \mathop{\mathbb{E}}_{y_t} [\hat{w}_{t,k} |_{y_t}] = 
\hat{c}_{t,k}/\sum_l \hat{c}_{t,l},
\end{align}
\fi
where $\hat{c} \equiv 1/\hat{\sigma}_{t,k}^2$, which can be viewed a count variable for the number of elements in each estimated cluster, subject to the decay factor $\rho$ over time.

\savespace{-0.1in}
\paragraph{Adding and removing clusters.} 
At any point in time, the mixture model is described by a collection of tuples
$(\hat{\rvp}_k, \hat{c}_k)$. We convert the probability of whether an
observation belongs to a new cluster into a decision: if $\hat{u}_t$ exceeds a
threshold $\alpha$, we create a new cluster. Due to the decay factor $\rho$,
our $\hat{c}$ estimate of a cluster can decay to zero over time, which is
appropriate for modeling nonstationary environments. In practice, we keep a
maximum number of $K$ clusters, and once the limit is reached, we simply pop
out the weakest  $\rvp_{k'}$, where $k'=\argmin(\hat{w}_k)$: $P_t = P_{t-1}
\setminus \{(\hat{\rvp}_{k'}, \hat{c}_{k'})\} \cup \{ (\rvz_t, 1) \}$.

\savespace{-0.1in}
\paragraph{Relation to Online ProtoNet.}
The formulation of our streaming EM-like algorithm is similar to the Online
ProtoNet~\citep{www}, with several key differences. First, to handle
nonstationary mixtures, we incorporate a decay term which is related to the
variance of the transition probability. Second, our new cluster creation is
unsupervised, whereas in~\citep{www}, only labeled examples lead to new
clusters. Third, representation learning in \citep{www} relies on a supervised
loss, whereas our objective---described in the next section---is entirely
unsupervised. Nonetheless, to indicate the lineage of our model, OUPN, we refer
to the cluster centroids as \emph{prototypes} and the mixture model as a
\emph{prototype memory}.

\savespace{-0.05in}
\subsection{Learning}
\label{sec:learning}
\savespace{-0.1in}
A primary goal of our learning algorithm is to learn good visual
representations through this online clustering process. We start the learning
from scratch: the encoder network is randomly initialized, and the prototype
memory will produce more accurate class predictions as the representations
become more informative throughout learning. Our overall representation
learning objective has three terms:
\ifarxiv
\begin{align}
    \gL &= \gL_\textrm{self} + \lambda_\textrm{ent} \gL_\textrm{ent} + \lambda_\textrm{new}
    \gL_\textrm{new}.
\end{align}
\fi
This loss function drives the learning of the main network parameters $\theta$,
as well as other learnable control parameters $\beta$, $\gamma$, and $\tau$. We
explain each term in detail below.

\begin{enumerate}[leftmargin=*]
\savespace{-0.05in}
\item \textbf{Self-supervised loss ($\gL_\textrm{self}$):} Inspired by recent
self-supervised representation learning approaches, we apply augmentations on
$\rvx_t$, and encourage the clustering assignments to match across different
views. Self-supervision follows three steps: First, the model makes a
prediction on the augmented view, and obtains $\hat{y}$ and $\hat{u}$ (E-step).
Secondly, it updates the prototype memory according to the prediction (M-step).
To create a learning target, we query the original view again, and obtain
$\tilde{y}$ to supervise the cluster assignment of the augmented view,
$\hat{y}'$, as in distillation~\citep{knowledgedistill}.
\ifarxiv
\begin{align}
\gL_\textrm{self} &= \frac{1}{T} \sum_t -\tilde{y}_t \log \hat{y}_t'.
\end{align}
\fi
Note that both $\tilde{y}_t$ and $\hat{y}_t'$ are produced after the M-step so
we can exclude the ``unknown'' class in the representation learning objective.
We here introduce a separate temperature parameter $\tilde{\tau}$ to control
the entropy of the mixture assignment $\tilde{y}_t$.

\savespace{-0.05in}
\item \textbf{Entropy loss ($\gL_\textrm{ent}$):} In order to encourage more
confident predictions we introduce a loss function $\gL_\textrm{ent}$ that
controls the entropy of the original prediction $\hat{y}$, produced in the
initial E-step:
\ifarxiv
\begin{align}
\gL_\textrm{ent} &= \frac{1}{T} \sum_t -\hat{y}_t \log \hat{y}_t.
\end{align}
\fi

\savespace{-0.05in}
\item \textbf{New cluster loss ($\gL_\textrm{new}$):} Lastly, our learning
formulation also includes a loss for initiating new clusters
$\gL_\textrm{new}$. We define it to be a Beta prior on the expected $\hat{u}$,
and we introduce a hyperparameter $\mu$ to control the expected number of
clusters:
\ifarxiv
\begin{align}
\gL_\textrm{new} &= -\log \Pr(\mathbbm{E}[\hat{u}]).
\end{align}
\fi
This acts as a regularizer on the total number of prototypes: if the system is
too aggressive in creating prototypes, then it does not learn to merge
instances of the same class; if it is too conservative, the representations can
collapse to a trivial solution.
\end{enumerate}
\savespace{-0.1in}
While there are several hyperparameters involved in inference and learning, in
our experiments we only optimize a few: the Beta mean $\mu$, the threshold
$\alpha$, the memory decay $\rho$, and the two loss term coefficients. The
others are set to default values for all datasets and experiments. See Appendix
\ref{sec:impl} for a complete discussion of hyperparameters.

\savespace{-0.1in}
\ifarxiv
\paragraph{Full algorithm.} 
Let $\Theta = \{\theta, \beta, \gamma, \tau\}$ denote the union of the
learnable parameters. Algorithm~\ref{alg:main} outlines our proposed learning
algorithm. The full list of hyperparameters are included in
Appendix~\ref{sec:impl}.
\ifarxiv
\else

\fi
\begin{minipage}{1.0\textwidth}
\ifarxiv
\else
\vspace{-0.2in}
\fi
\begin{algorithm}[H]
\caption{Online Unsupervised Prototypical Learning}
\label{alg:main}
\begin{algorithmic}
\REPEAT
\STATE $\gL_{\textrm{self}} \gets 0$, $p_{\textrm{new}} \gets 0.$
\FOR {$t \gets 1 \dots T$}
\STATE Observe new input $\rvx_t$.
\STATE Encode input, $\rvz_t \gets h(\rvx_t; \theta)$.
\STATE Compare to existing prototypes: $[\hat{u}_t, \hat{y}_t] \gets \textrm{E-step}(\rvz_t, P; \beta, \gamma, \tau).$
\IF{$\hat{u}^0_t < \alpha$}
    \STATE Assign $\rvz_t$ to existing prototypes: $P \gets \textrm{M-step}(\rvz_t, P, \hat{u}_t, \hat{y}_t)$.
\ELSE
    \STATE Recycle the least used prototype if $P$ is full.
    \STATE Create a new prototype $P \gets P \cup \{(\rvz_t, 1)\}$.
\ENDIF
\STATE Compute pseudo-labels: $[\_, \tilde{y}_t] \gets \textrm{E-step}(\rvz_t, P; \beta, \gamma, \tilde{\tau}).$
\STATE Augment a view: $\rvx_t' \gets \textrm{augment}(\rvx_t).$
\STATE Encode the augmented view: ${\rvz}_t' \gets h(\rvx_t'; \theta).$
\STATE Compare the augmented view to existing prototypes: $[\_, \hat{y}_t'] \gets \textrm{E-step}(\rvz_t', P; \beta, \gamma, \tau).$
\STATE Compute the self-supervision loss: $\gL_{\textrm{self}} \gets \gL_{\textrm{self}} - \frac{1}{T} \tilde{y}_t \log \hat{y}_{t}'.$
\STATE Compute the entropy loss:  $\gL_{\textrm{ent}} \gets \gL_{\textrm{ent}} - \frac{1}{T} \hat{y}_t \log \hat{y}_t.$
\STATE Compute the average probability of creating new prototypes, $p_{\textrm{new}} \gets p_{\textrm{new}} + \frac{1}{T}\hat{u}_t.$
\ENDFOR
\STATE Compute the new cluster loss: $\gL_{\textrm{new}} \gets -\log\Pr(p_{\textrm{new}}).$
\STATE Sum up losses: $\gL \gets \gL_{\textrm{self}} + \lambda_{\textrm{ent}} \gL_{\textrm{ent}} + \lambda_{\textrm{new}} \gL_{\textrm{new}}.$
\STATE Update parameters: $\Theta \gets \textrm{optimize}(\gL, \Theta).$
\UNTIL{convergence}
\RETURN $\Theta$
\end{algorithmic}
\end{algorithm}
\end{minipage}

It is worth noting that if we create a new prototype every time step, then OUPN
is similar to a standard contrastive learning with an instance-based InfoNCE
loss~\citep{simclr,moco}; therefore it can be viewed as a generalization of
this approach. Additionally, all the losses can be computed online without
having to store any examples beyond the collection of prototypes.
\fi

%% file: sections/experiments.tex
\section{Experiments}
\savespace{-0.1in}

\begin{figure*}[t]
\savespacefigtop{-0.5in}
\centering
\includegraphics[width=\textwidth,trim={0, 5.5cm, 1.8cm,
0},clip]{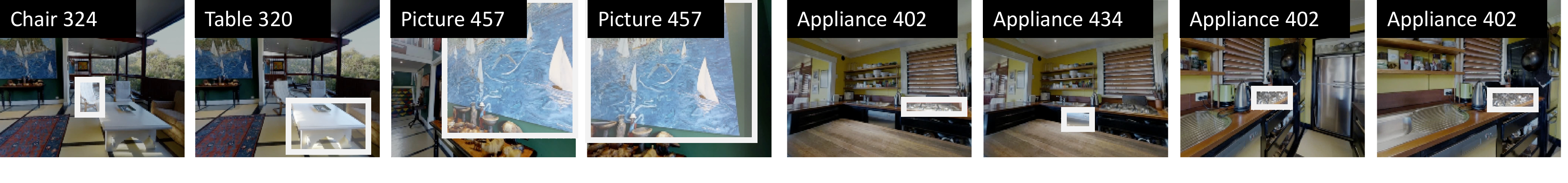}
\savespace{-0.3in}
\morespace{-0.2in}
\caption{An example subsequence of the \rro{} dataset~\citep{www}, consisting
of glimpses of an agent roaming in an indoor environment, and the task is to
recognize object instances.}
\label{fig:roamingrooms}
\savespace{-0.15in}
\end{figure*}

In this section, we evaluate our proposed learning algorithm on a set of visual
learning tasks and examine the quality of the output categories. Contrasting 
with prior work on visual representation learning, our primary scenario of
interest is online training with non-iid image sequences.

\savespace{-0.1in}
\paragraph{Implementation details.} 
\looseness=-1
Throughout our experiments, we make two changes to the model inference
procedure defined above. First, we use cosine similarity instead of negative
squared Euclidean distance for computing the mixture logits, because cosine
similarity is bounded and is found to be more stable to train. Second, when we
perform cluster inference, we treat the mixing coefficients $w_k$ as constant
and uniform as otherwise we find that the representations may collapse into a
single large cluster.
\ifarxiv
\footnote{Our source code is released at:
\url{https://github.com/renmengye/online-unsup-proto-net}.}
\else
\fi

\savespace{-0.1in}
\paragraph{Online clustering evaluation.} 
During evaluation we present our model a sequence of all new images (unlabeled
or labeled) and we would like to see how well it produces a successful grouping
of novel inputs. The class label index starts from zero for each sequence, and
the classes  do not overlap with the training set. The model memory is reset at
the beginning of each sequence.

In unsupervised readout, the model directly predicts the label for each image,
i.e. the model $g$ directly predicts $\hat{y}_t = g(\rvx_{1:t})$. In supervised
readout (\emph{for evaluation only}), the model has access to all labels up to
time step $t-1$, and needs to predict the label for the $t$-th image, i.e.
$\hat{y}_t = g(\rvx_{1:t}, y_{1:t-1})$. We used the following metrics to
evaluate the quality of the grouping of test sequences:

\begin{itemize}[leftmargin=*]
\savespace{-0.1in}
\item 
\textbf{Adjusted mutual information (AMI):} In the \textit{unsupervised}
setting, we use the mutual information metric to evaluate the similarity
between our prediction $\{\hat{y}_1, \dots,
\hat{y}_T\}$ the groundtruth class ID $\{y_1, \dots, y_T\}$. Since the online
greedy clustering method admits a threshold parameter $\alpha$ to control the
number of output clusters, therefore for each model we sweep the value of
$\alpha$ to maximize the AMI score, to make the score threshold-invariant:
$
\textrm{AMI}_{\max} = \max_\alpha \textrm{AMI}(y, \hat{y}(\alpha)).
$
The maximization of $\alpha$ can be thought of as part of the readout
procedure, and it is designed to particularly help other self-supervised
learning baselines since their feature similarity functions are not necessarily
calibrated for clustering.
\savespace{-0.05in}
\item \textbf{Average precision (AP):} In the \textit{supervised} setting, we
followed the evaluation procedure in \citet{www} and used average precision,
which combines both accuracy for predicting known classes as well as unknown
ones.
\end{itemize}

\savespace{-0.15in}
\paragraph{Offline readout evaluation.} 
\looseness=-100000
A popular protocol to evaluate self-supervised representation learning is to
use a classifier trained offline on top of the representations to perform
semantic class readout. Because AMI and AP are designed to evaluate novel
instance classification, we included offline evaluation protocols for semantic
classes. We considered the following classifiers:
\begin{itemize}[leftmargin=*]
\savespace{-0.1in}
\item \textbf{Nearest neighbor readout:} A common protocol is to use a
k-nearest-neighbor classifier to readout the learned representations. For
\rro{} we set $k=39$ and for SAYCam we set $k=1$.
\item \textbf{Linear readout:} Another popular protocol is to train a linear
classifier on top of the learned representations to a given set of semantic
classes. For \rro{}, we used the Adam optimizer with learning rate $10^{-3}$
for 20 epochs, and for SAYCam, we used the SGD optimizer with learning rate
searched among \{1.0, 0.1, 0.01\} for each model for 100 epochs.
\end{itemize}

\begin{figure*}[t]
\savespacefigtop{-0.5in}
\centering
\includegraphics[width=\textwidth]{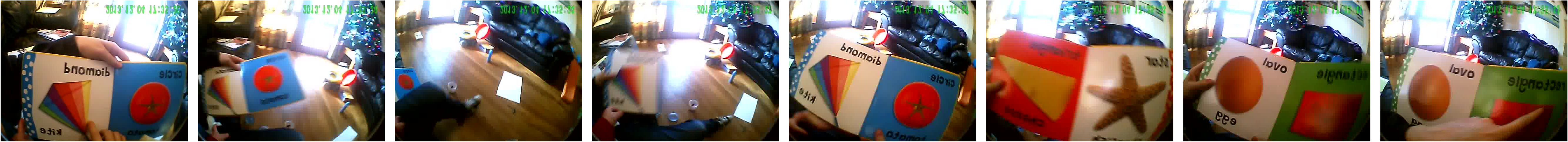}
\savespace{-0.2in}
\morespace{-0.2in}
\caption{An example subsequence of the SAYCam dataset~\citep{saycam},
consisting of egocentric videos collected from human babies.}
\label{fig:babycam}
\savespace{-0.2in}
\end{figure*}

\begin{table}[t]
\savespacefigtop{-0.9in}
\begin{minipage}[b]{.55\linewidth}
\begin{center}
\begin{small}
\savespace{-0.1in}
\resizebox{!}{2.3cm}{
\begin{tabular}{lcccc}
\toprule
                            & \multirow{2}{*}{AMI} & \multirow{2}{*}{AP} & Acc. & Acc.             \\
                            &       &            & (k-NN,\%)       & (Linear,\%)      \\
\midrule                                                                               
\multicolumn{5}{l}{\bf Supervised}                                                   \\
\midrule                                                                               
Supervised CNN              & -       & -        &  72.11         & 71.93            \\
Online ProtoNet~\citep{www} & 79.02   & 89.94    &  -             & -                \\
\midrule                                                                               
\multicolumn{5}{l}{\bf Unsupervised}                                                 \\
\midrule                                                                               
Random Network              & 28.25    & 11.68    & 28.84         & 26.73            \\
SimCLR~\citep{simclr}       & 50.03    & 52.98    & 44.84         & 48.83            \\
SwAV~\citep{swav}           & 42.70    & 37.31    & 40.04         & 45.77            \\
SwAV+Queue~\citep{swav}     & 48.31    & 50.40    & 43.63         & 45.31            \\
SimSiam~\citep{simsiam}     & 47.58    & 44.15    & 43.99         & 48.24            \\
OUPN (Ours)                 &\bf{78.16}&\bf{84.86}& \bf{48.37}    & \bf{52.28}       \\
\bottomrule
\end{tabular}
}
\end{small}
\end{center}
\caption{Instance and semantic class recognition results on \rro{}}
\label{tab:rooms}
\end{minipage}
\hfill
\begin{minipage}[b]{.4\linewidth}
\begin{center}
\begin{small}
\resizebox{!}{2.7cm}{
\begin{tabular}{lccc}
\toprule
Random split & Acc. 1-NN       & Linear   \\
\midrule
ImageNet      &  72.67     &  53.23  \\
TC-S~\citep{eyesofachild} (iid) &  80.76 & 62.36    \\
\midrule
Random	& 10.04 & 9.37  \\
SimSiam	& 26.53 & 19.91 \\
SwAV	& 34.48 & 31.99 \\
SimCLR	& 49.13 & 37.23 \\
OUPN (Ours)	& \textbf{64.35} & \textbf{44.29} \\
\midrule
Subsample 10x  & Acc. 1-NN       & Linear   \\
\midrule
ImageNet      &   48.09    &  40.61  \\
TC-S~\citep{eyesofachild} (iid) & 60.43   & 50.16      \\
\midrule
Random	    & 9.74   & 15.15 \\
SimSiam	    & 18.71  & 17.39 \\
SwAV	    & 21.90  & 19.89 \\
SimCLR	    & 28.24  & 25.98 \\
OUPN (Ours)	& \textbf{36.52}  & \textbf{30.25} \\
\bottomrule
\end{tabular}
}
\end{small}
\end{center}
\caption{Semantic classification results on SAYCam (Child S)}
\label{tab:saycam}
\end{minipage}
\end{table}

\savespace{-0.1in}
\paragraph{Competitive methods.}
\looseness=-10000
Our focus is online unsupervised visual representation learning. There are very
few existing methods developed for this setting. To the best of our knowledge,
continual unsupervised learning~\citep{curl} (CURL) is the only directly
comparable work, but this method relies on input reconstruction and scales
poorly to more general environments. We include the comparison to CURL in the
Appendix (Table~\ref{tab:iidmlp}). Unsupervised few-shot learning approaches
are also related~\citep{umtra,protoclr}, but these methods are directly related
to standard self-supervised learning methods. Therefore we compare OUPN with
the following competitive self-supervised visual representation learning
methods.

\begin{itemize}[leftmargin=*]
    \savespace{-0.1in}
    \looseness=-1
    \item \textbf{SimCLR}~\citep{simclr} is a contrastive learning method with
    an instance-based objective that tries to classify an image instance among
    other augmented views of the same batch of instances. It relies on a large
    batch size and is often trained on well-curated datasets such as
    ImageNet~\citep{imagenet}.
    
    \savespace{-0.05in}
    \looseness=-1
    \item \textbf{SwAV}~\citep{swav} is a contrastive learning method with a
    clustering-based objective. It has a stronger performance than SimCLR on
    ImageNet. The clustering is achieved through Sinkhorn-Knopp which assumes
    balanced assignment, and prototypes are learned by gradient descent.
    \savespace{-0.05in}
    \item \textbf{SwAV+Queue} is a SwAV variant with an additional example
    queue. This setup is proposed in~\citet{swav} to deal with small training
    batches. A feature queue that accumulates instances across batches allows
    the clustering process to access more data points. The queue size is set to
    2000.
    \savespace{-0.05in}
    \item \textbf{SimSiam}~\citep{simsiam} is a self-supervised
    learning method that does not require negative samples. It uses a
    stop-gradient mechanism and a predictor network to make sure the
    representations do not collapse. Through not using negative samples,
    SimSiam could be immune to treating images of the same instances as
    negative samples.
\end{itemize}
\savespace{-0.1in}
For fair comparison on online representation learning, all of the above methods
are trained on the \emph{same} dataset using the same input data as our model,
instead of using their pretrained checkpoints from ImageNet.

Since none of these competitive methods are designed to output classes with a
few examples, we need an additional clustering-based readout procedure to
compute AMI and AP scores. We use a simple online greedy clustering procedure
for these methods. For each timestep, it searches for the closest prototype; in
unsupervised mode, if it fails with $\hat{u}_t$ greater than $\alpha$, it will
create a new prototype, and otherwise it will aggregate the current embedding
to the cluster centroid. As explained above, the $\alpha$ parameter is
maximized for each model on its test scores to optimize performance.

\begin{figure*}[t]
\centering
\includegraphics[height=1.5cm,trim={1cm 17cm 0.5cm 3.3cm},clip]{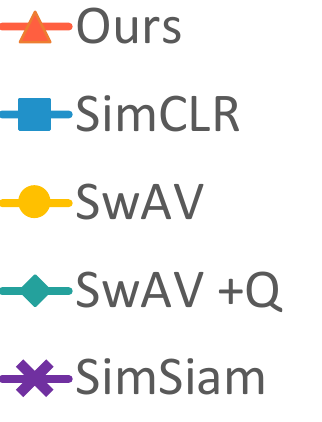}
\includegraphics[height=3.7cm,trim={0.1cm 0.1cm 0.1cm 0.1cm},clip]{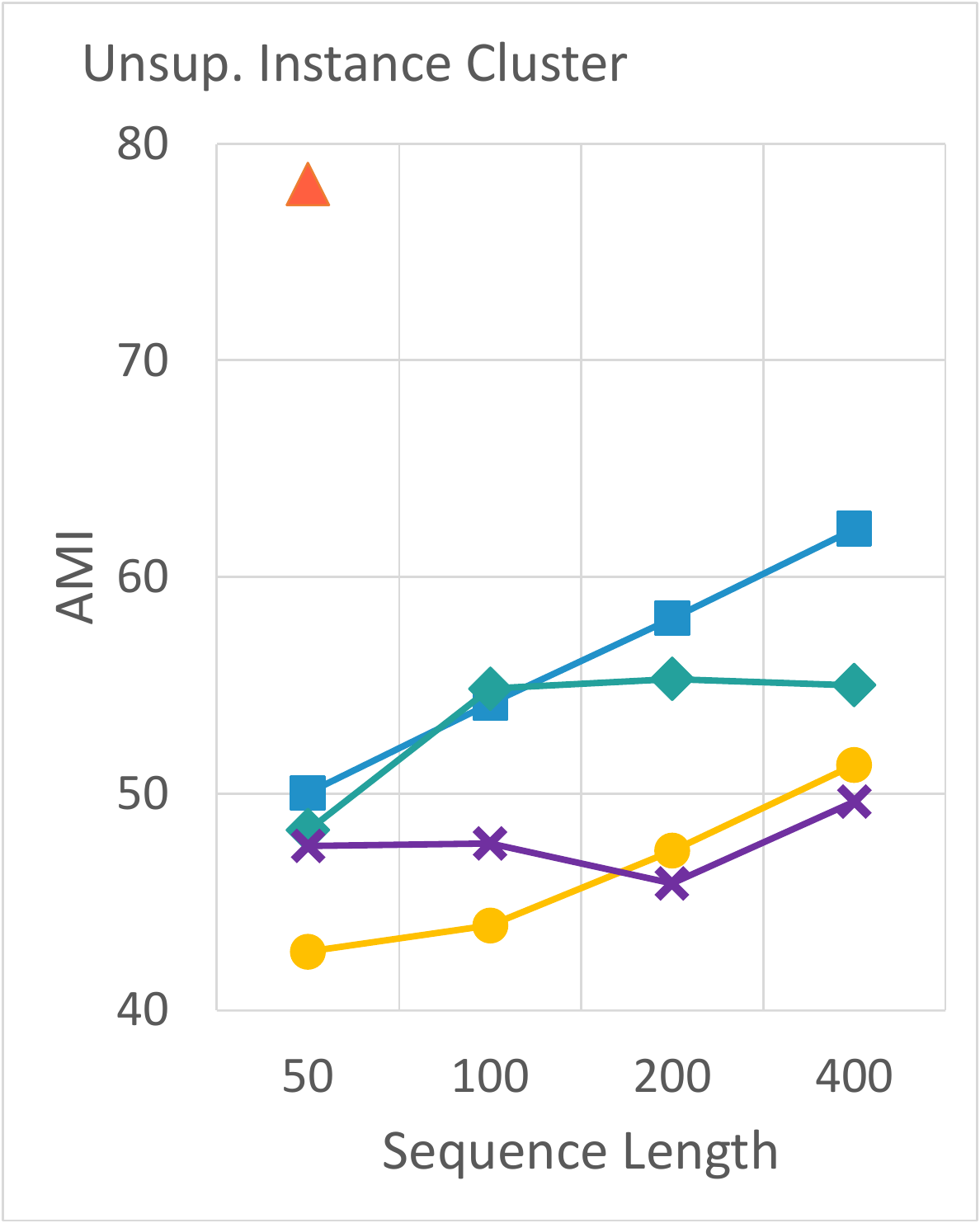}
\includegraphics[height=3.7cm,trim={0.1cm 0.1cm 0.1cm 0.1cm},clip]{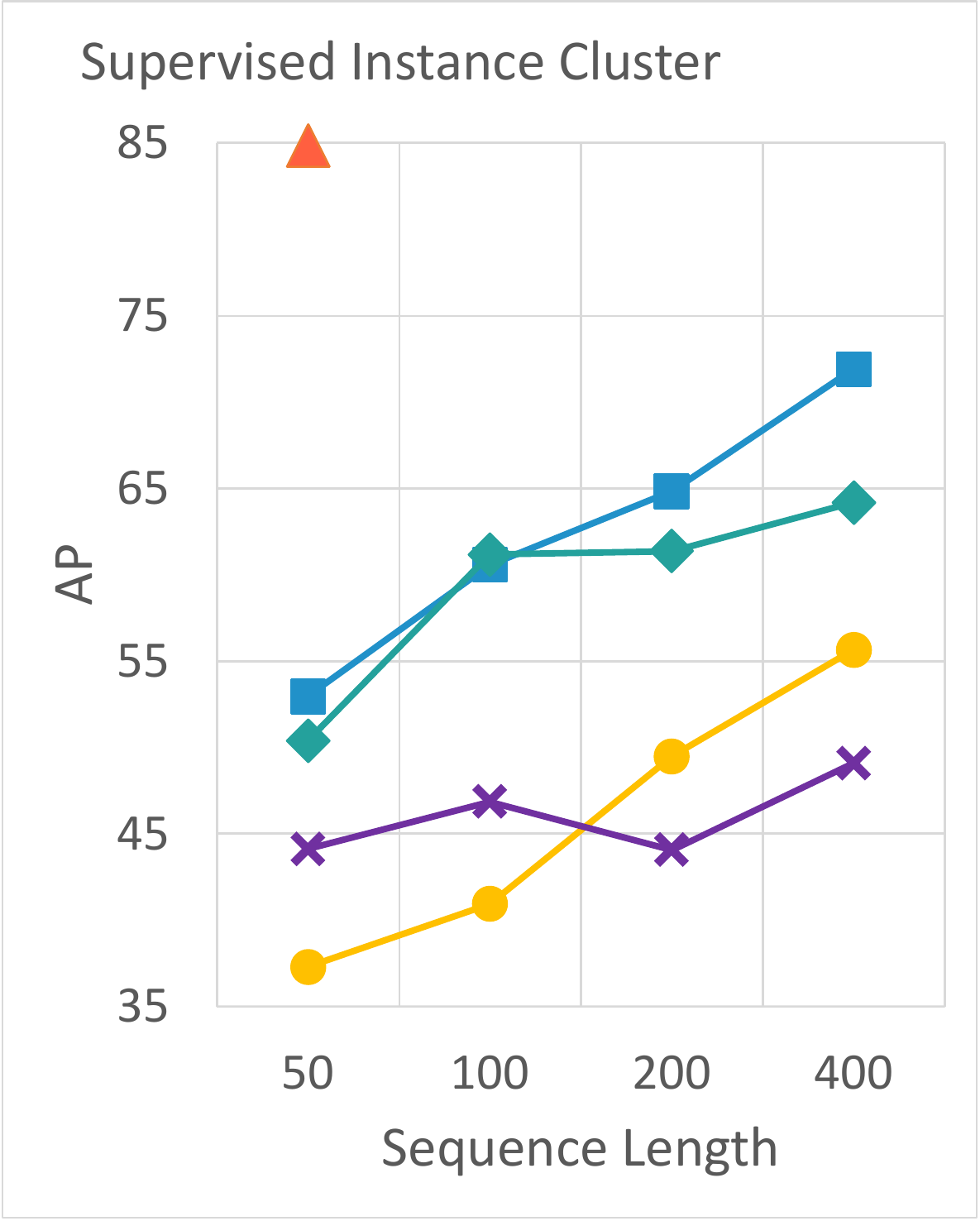}
\includegraphics[height=3.7cm,trim={0.1cm 0.1cm 0.1cm 0.1cm},clip]{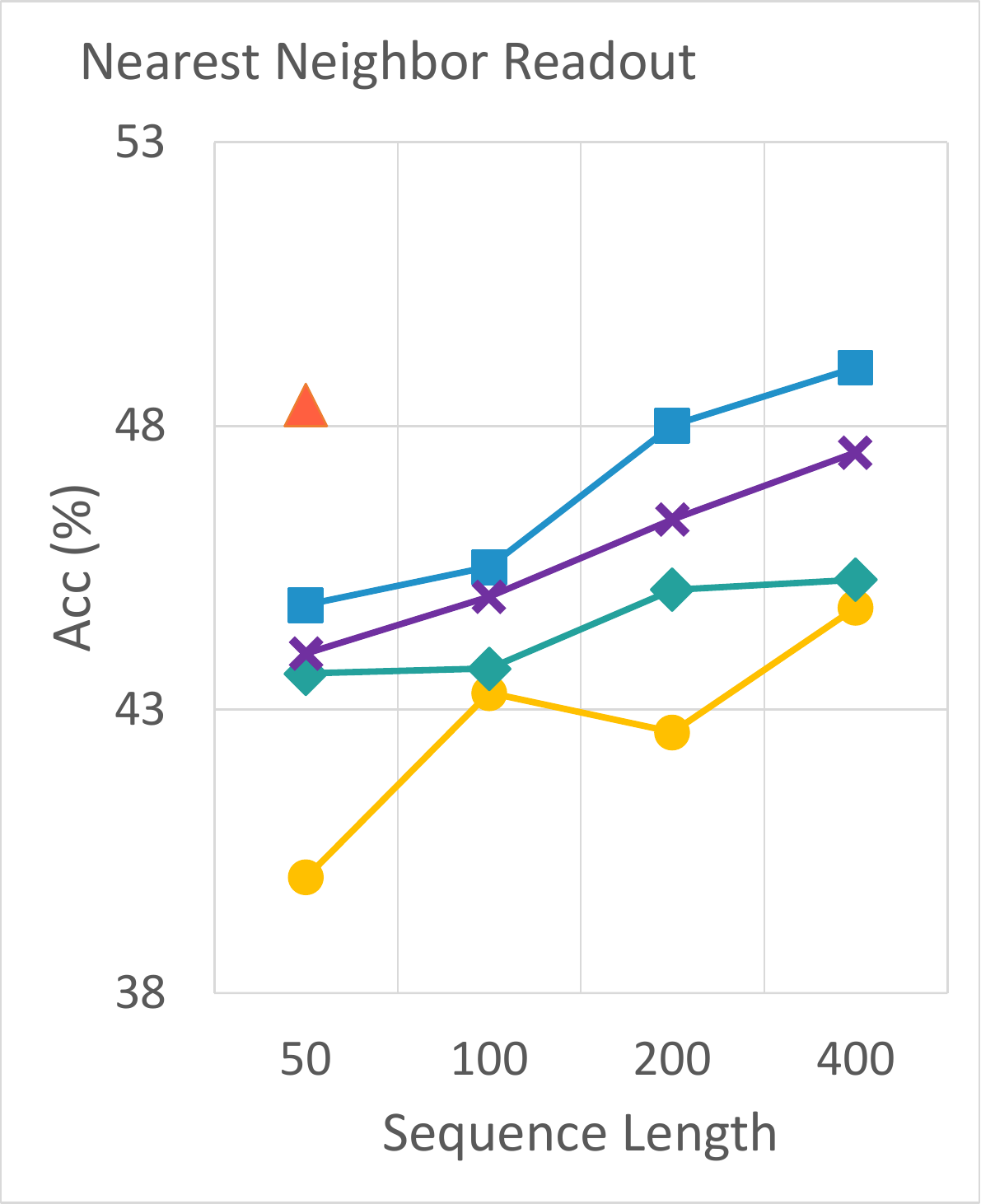}
\includegraphics[height=3.7cm,trim={0.1cm 0.1cm 0.1cm 0.1cm},clip]{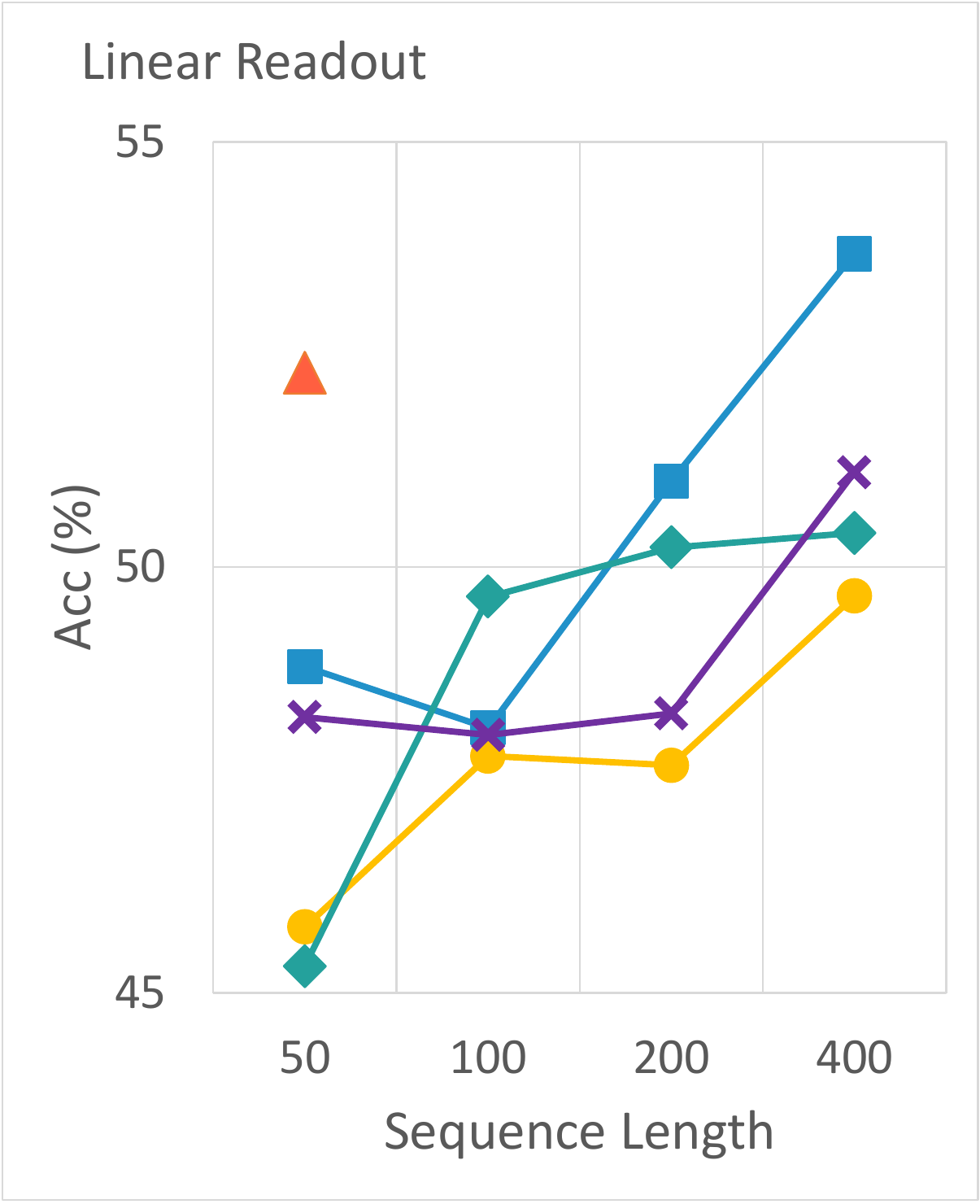}
\savespace{-0.05in}
\caption{Comparison to SimCLR, SwAV, and SimSiam with larger batch sizes on \rro{}}
\label{fig:batchsize}
\savespace{-0.15in}
\end{figure*}

\savespace{-0.1in}
\subsection{Indoor home environments}
\savespace{-0.1in}
\looseness=-10000
We first evaluate the algorithm using the \rro{} dataset~\citep{www} where the
images are collected from indoor environments~\citep{matterport} using a random
walking agent. The dataset contains 1.22M image frames and 7K instance classes
from 6.9K random walk episodes.
\ifarxiv
Each image is resized to 120 $\times$ 160 $\times$ 3. We use a maximum of 50
frames for training due to memory constraints, and all methods are evaluated on
the test set with a maximum of 100 frames per episode.
\fi
The input for each frame is the RGB values and a object segmentation mask (in
the 4th channel); the output (used here only for evaluation with AP and AMI) is
the object instance ID. An example episode is shown in
Fig.~\ref{fig:roamingrooms}. The dataset is split into different home
environments (60 training, 10 val, and 20 test). Each training iteration
consists of a sequence of images from one of the homes. At test time, for the
instance classification task, we ask the model to recognize novel objects in a
new sequence of images in one of the test homes. For the semantic
classification task, we ask the model to classify among 21 semantic categories
including ``picture'', ``chair'', ``lighting'', ``cushion'', ``table'', etc.

SimCLR, SwAV and SimSiam use varying batch sizes (50, 100, 200, and 400). For
online (non-iid) settings, the notion of batch size can be understood as
``sequence length''. Other training parameters can be found in the Appendix.
Note that all baselines use the same training inputs as our model.

\ifarxiv
\savespace{-0.1in}
\paragraph{Implementation details.} 
\looseness=-1
We use a ResNet-12~\citep{tadam} as the encoder network, and we train our
method over 80k 50-frame episodes (4M image frames total), using a single
NVIDIA 1080-Ti GPU. We follow the same procedure of image augmentation as
SimCLR~\citep{simclr}. We use 150 prototypes with $\rho=0.995$. More
implementation details can be found in Appendix~\ref{sec:impl}.
\else
\fi

\savespace{-0.1in}
\paragraph{Results.} Our main results are shown in Table~\ref{tab:rooms}.
Although self-supervised methods, such as SimCLR, SwAV and SimSiam, have shown
promising results on large batch learning on ImageNet, their performance here
are relatively weak compared to the supervised baseline. In contrast, our
method OUPN shows impressive performance on this benchmark: it almost matches
the supervised learner in AMI, and reached almost 95\% of the performance of
the supervised learner in AP. OUPN also outperforms other methods in terms of
k-NN and linear readout accuracy. We hypothesize that the nonstationary
distribution of online frames could impose several challenges to standard
self-supervised learning methods. First, SimCLR could treat adjacent similar
frames as negative pairs. Second, it breaks SwAV's assumption on balanced
cluster assignment and stationary cluster centroids. Adding a queue slightly
improves SwAV; however, since the examples in the queue cannot be used to
compute gradients, the nonstationary distribution still hampers gradient
updates. Lastly, all of them could suffer from a very small batch size in our
online setting.

To illustrate the impact of our small batch episodes, we increase the batch
size for SimCLR and SwAV, from 50 to 400, at the cost of using multiple GPUs
training in parallel. The results are shown in Fig.~\ref{fig:batchsize}.
Results indicate that increasing the batch size can improve these baselines,
which matches our expectation. Nevertheless, our method using a batch size of
50 is still able to outperform other self-supervised methods using a batch size
of 400, which takes 8$\times$ computational resource compared to ours. Note
that the large batch experiments are designed to provide the best setting for
other self-supervised methods to succeed. We do not need to run our model with
larger batch size since our prototype memory is a sequential module, and
keeping the batch size smaller allows quicker online adaptation and less memory
consumption.

\begin{figure*}[t]
\savespacefigtop{-0.5in}
\centering
\includegraphics[height=1.5cm,trim={1cm 17cm 0.5cm 3.3cm},clip]{figures/legend.pdf}
\includegraphics[height=3.7cm,trim={0.1cm 0.1cm 0.1cm 0.1cm},clip]{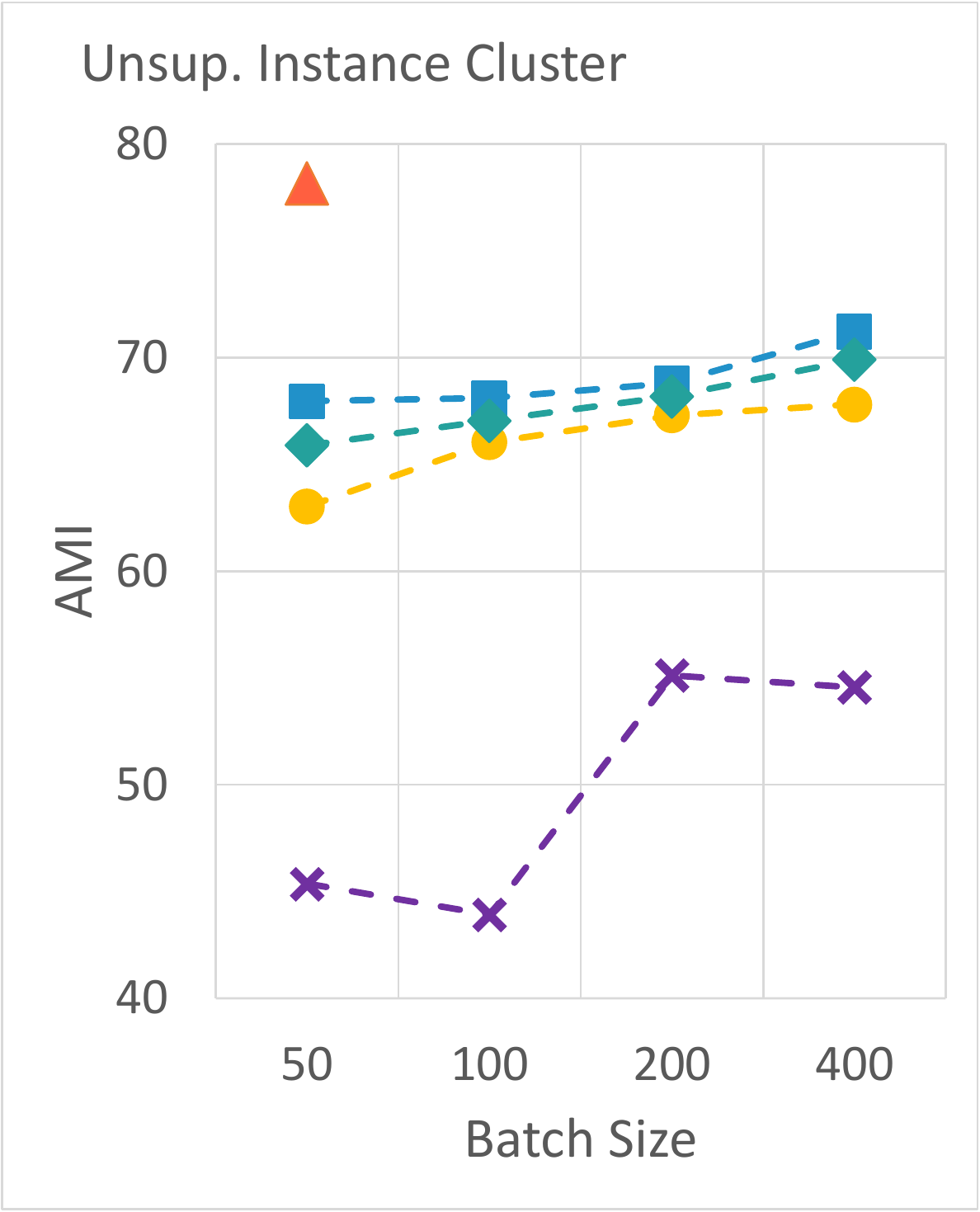}
\includegraphics[height=3.7cm,trim={0.1cm 0.1cm 0.1cm 0.1cm},clip]{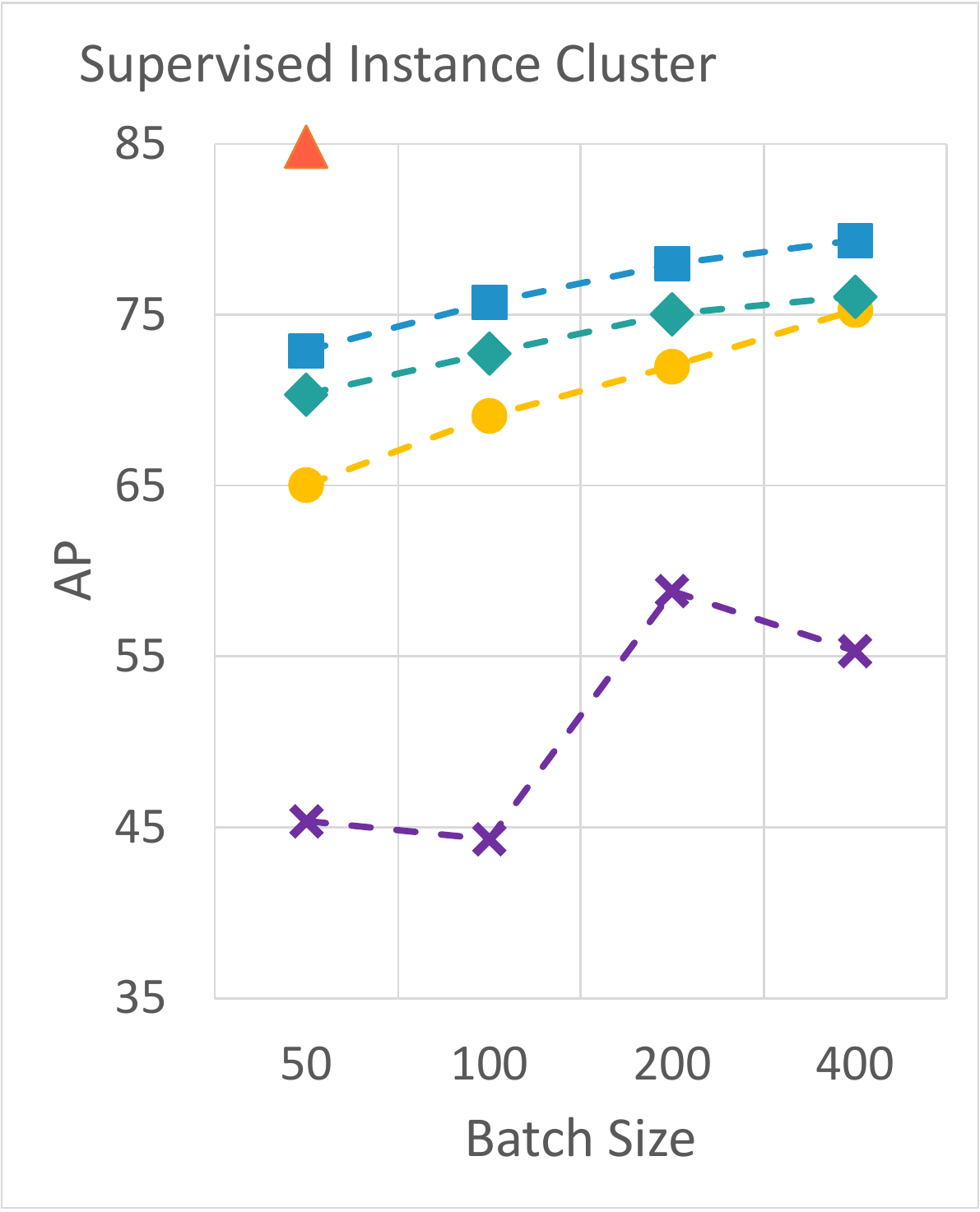}
\includegraphics[height=3.7cm,trim={0.1cm 0.1cm 0.1cm 0.1cm},clip]{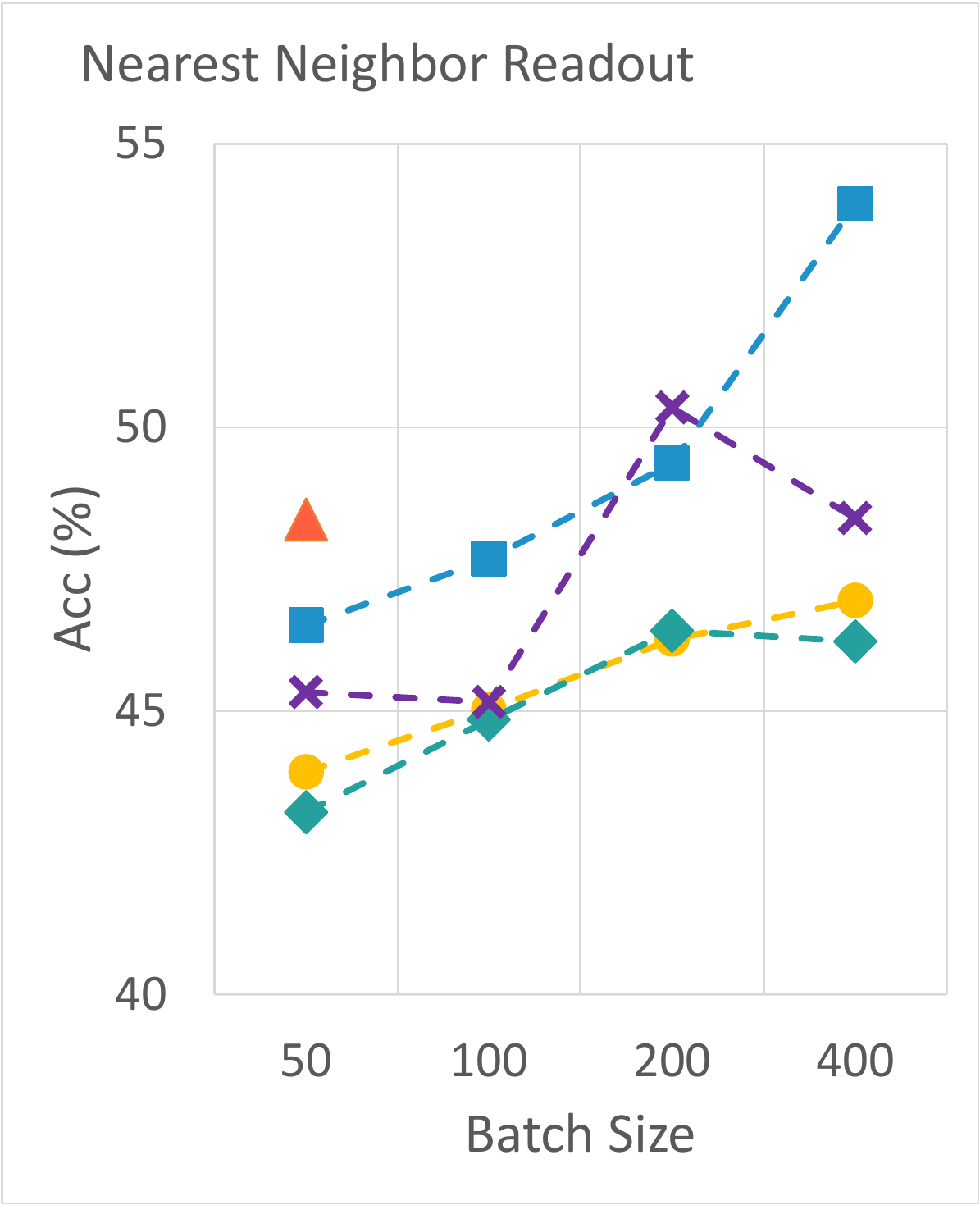}
\includegraphics[height=3.7cm,trim={0.1cm 0.1cm 0.1cm 0.1cm},clip]{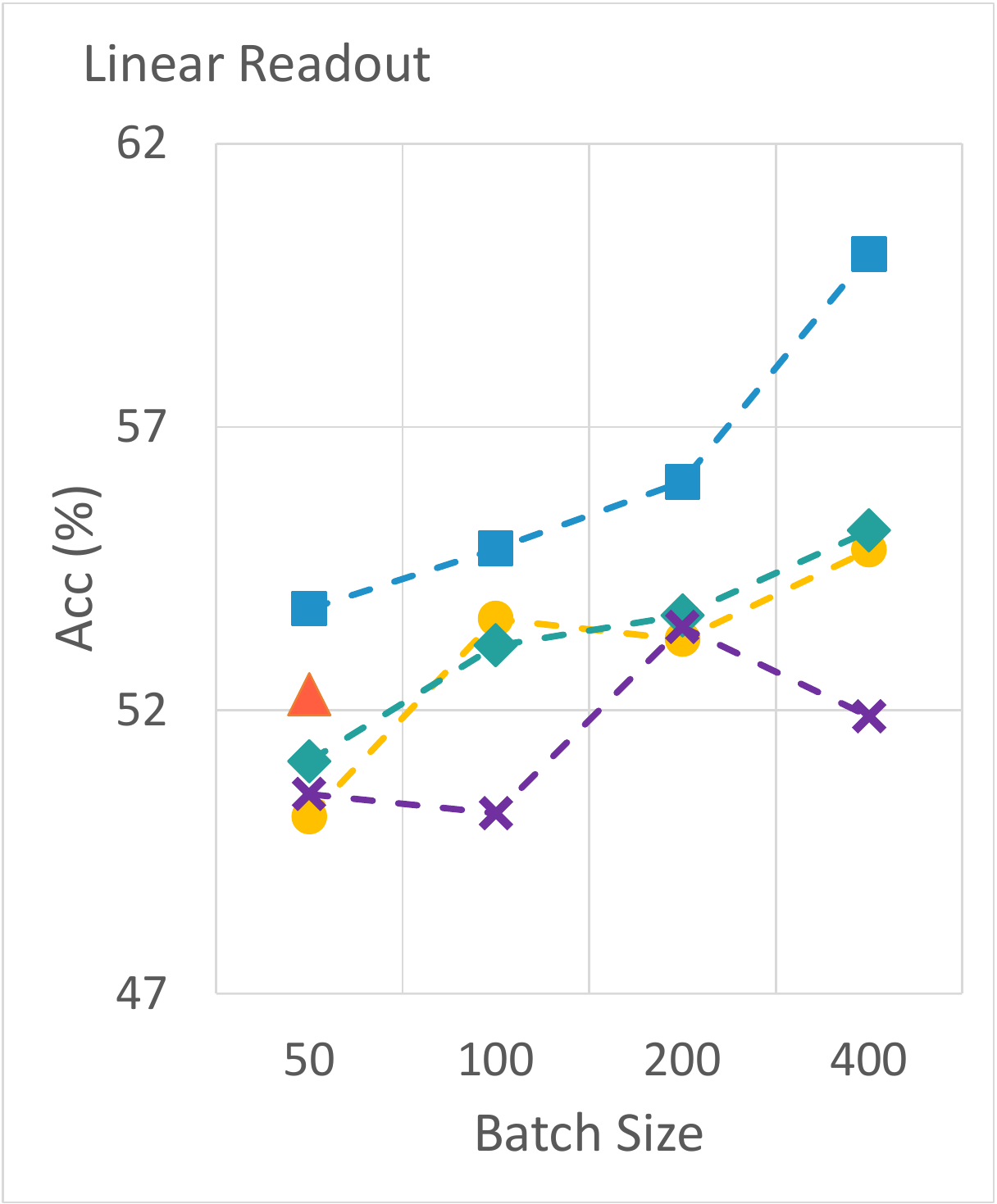}
\savespace{-0.05in}
\caption{Comparison to iid-trained versions of SimCLR, SwAV, and SimSiam on \rro{}.}
\label{fig:iid}
\savespace{-0.15in}
\end{figure*}

\savespace{-0.15in}
\paragraph{Comparison to iid modes of SimCLR, SwAV, and SimSiam.} 
\looseness=-10000
The original SimCLR, SwAV, and SimSiam were designed to train on iid data. To
study the effects of this assumption, we implemented an approximation to an iid
distribution by using a large random queue that shuffles the image frames. As
in the study shown in Fig.~\ref{fig:iid}, we again vary the batch size for
these competitive methods. All of these self-supervised baselines thrive with
iid data; the gains of iid over non-iid can be seen by comparing
Fig.~\ref{fig:iid} to Fig.~\ref{fig:batchsize}. Larger batches help both
methods again here. Interestingly, our method using a batch size of 50 non-iid
data again outperforms both methods using a batch size of 400 of iid data in
terms of AMI and AP. The only case where our method is inferior to SimCLR is
when SimCLR is trained with large batches under iid setting on semantic
classification readout. This is reasonable since semantic classification and
iid large batch training is the setting SimCLR was originally developed for.
Again, iid large batch training is not what we aim to solve in this paper, and
we include the iid experiments in the paper simply to better understand the
failure points of existing algorithms.

\begin{figure*}[t]
\savespacefigtop{-0.5in}
\centering
\ifarxiv
\includegraphics[width=\textwidth,trim={0, 4.0cm, 0, 0cm},clip]{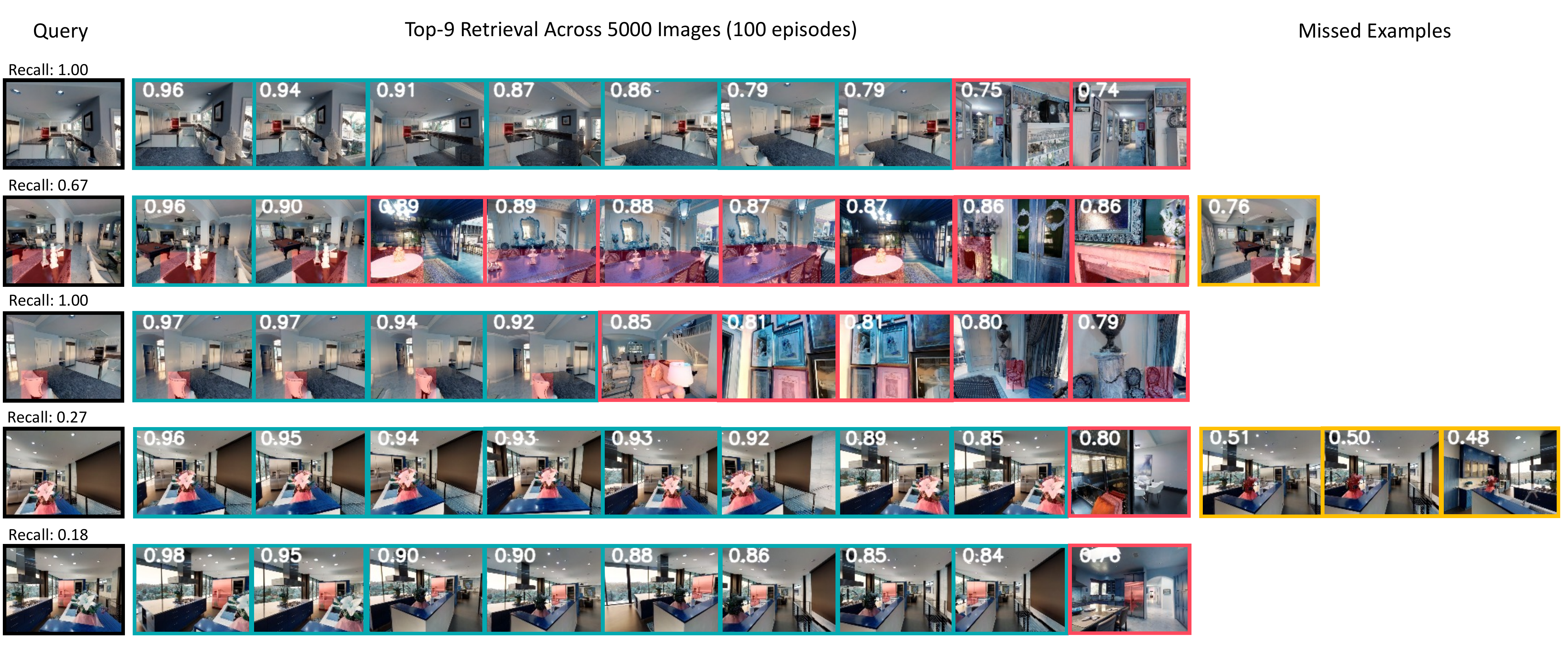}
\fi
\caption{Image retrieval results on \rro{}. In each row, the leftmost image is
the query image, and retrieved images are shown to its right. Cosine similarity
scores on the top left; a green border denotes a correct retrieval, red false
positive, and yellow a miss. Recall is the proportion of instances in the
top-9.}
\label{fig:retrieval}
\savespace{-0.15in}
\end{figure*}

\savespace{-0.15in}
\paragraph{Visualization on image retrieval.} 
To verify the usefulness of the learned representation, we ran an image
retrieval visualization using the first 5000 images in the first 100 test
sequences of length 50 and perform retrieval in a leave-one-out procedure. This
procedure is only to visualize the similarity and is distinct from our
evaluation procedure that requires class-label prediction.The results are shown
in Fig.~\ref{fig:retrieval}. Similarity scores are also provided. The top
retrieved images are all from the same instance of the query image, and our
model sometimes achieves perfect recall. This confirms that our model can
handle a certain amount of view angle change. We also investigated the missed
examples and we found that these are taken from more distinct view angles.

\savespace{-0.1in}
\subsection{Head mounted camera recordings}
\label{sec:saycam}
\savespace{-0.1in}
\looseness=-10000
Inspired by how humans acquire visual understanding ability after birth, we
further evaluated our method on realistic first-person videos collected using
baby egocentric cameras. The SAYCam dataset~\citep{saycam} is collected using
500 hours video data from three children. We obtained permission to use from
the original authors. Following prior work~\citep{eyesofachild}, we focused on
using the Child S subset in our work. See Figure~\ref{fig:babycam} for an
example subsequence. We used MobileNet-V2 for this experiment. We sampled the
video at 4 seconds per frame to form a temporal window of 5 minutes (75 images)
for each mini-batch. The inputs are cropped and reshaped into 224 $\times$ 224
RGB images. We repeat the full 164-hour video for 16 times (16 epochs) for a
total of 2624 hours for all methods trained on this dataset.

\looseness=-10000
To evaluate the learned representations, \citet{eyesofachild} used a labeled
dataset of the images containing 26 semantic classes such as \textit{ball},
\textit{basket}, \textit{car}, \textit{chair}, etc. Following their settings,
we used two different splits of the dataset: a random iid split and a
subsampled split, which was proposed to reduce the proportion of redundant
images. We used both a linear and a nearest neighbor readout.

\savespace{-0.15in}
\paragraph{Results.} 
\looseness=-10000
Results are shown in Table~\ref{tab:saycam}. We are able to outperform
competitive self-supervised learning methods. We also reproduced the
performance of the temporal classification (TC) model~\citep{eyesofachild} and
an ImageNet pretrained model for comparison. Since the TC model is trained
using random iid samples of the full video, therefore it is understandable that
our online streaming model performs worse. We also note that nearest neighbor
readout generally performs better than linear readout on this benchmark, likely
due to the existence of many similar frames in the video.

\savespace{-0.1in}
\subsection{Handwritten characters and ImageNet images}
\savespace{-0.1in}
\looseness=-1000
We also evaluated our method on two different tasks: recognizing novel
handwritten characters from Omniglot~\citep{lake2015} and novel ImageNet
classes. Here, images are static and are not organized in a video-like
sequence, and models have to reason more about conceptual similarity between
images to learn grouping. Furthermore, since this is a more controllable setup,
we can test our hypothesis concerning sensitivity to class imbalance by
performing manipulations on the episode distribution.

\begin{wraptable}{r}{0.53\textwidth}
\centering
\vspace{-0.2in}
\resizebox{!}{1.9cm}{
\begin{tabular}{lcccccc}
\toprule
& \multicolumn{2}{c}{\rom{}} & \multicolumn{2}{c}{\rim{}} \\
                             & AMI       & AP         & AMI       & AP        \\
\midrule                                               
\multicolumn{3}{l}{\bf Supervised}                   \\
\midrule                                               
Pretrain-Supervised         & 84.48      & 93.83   & 29.44     & 24.39       \\
Online ProtoNet~\citep{www} & 89.64      & 92.58   & 29.73     & 25.38     \\
\midrule                                               
\multicolumn{3}{l}{\bf Unsupervised}                 \\
\midrule                                               
Random Network              & 17.66      & 17.01   & 4.55      & 2.65      \\
SimCLR~\citep{simclr}       & 59.06      & 73.50   & 6.87      & 12.25     \\
SwAV~\citep{swav}           & 62.09      & 75.93   & 9.87      & 5.23      \\
SwAV+Queue~\citep{swav}     & 67.25      & 81.96   & 10.61     & 4.83      \\
SimSiam~\citep{simsiam}     & 45.57      & 56.12    & 12.64     & 6.31      \\ 
OUPN (Ours)                 & \bf{84.42} &\bf{92.84} & \bf 19.03 & \bf 15.05 \\
\bottomrule
\end{tabular}
}
\savespace{-0.05in}
\ifarxiv
\caption{\small \rom{} and \rim{}}
\fi
\label{tab:omniglot}
\savespace{-0.2in}
\morespace{-0.2in}
\end{wraptable}

Our episodes are sampled from the \rom{} and \rim{} dataset~\citep{www}. An
episode involves several different \emph{contexts}, each consisting of a set of
classes, and in each context, classes are sampled from a Chinese restaurant
process. We use 150-frame episodes with 5 contexts for \rom{} and 48-frame with
3 contexts for \rim{}.

\begin{figure}[t]
\centering
\savespacefigtop{-0.5in}
\begin{minipage}[c]{0.6\linewidth}
\centering
\ifarxiv
\includegraphics[height=3.6cm,trim={0.615cm 0.1cm 0.33cm 0.1cm},clip]{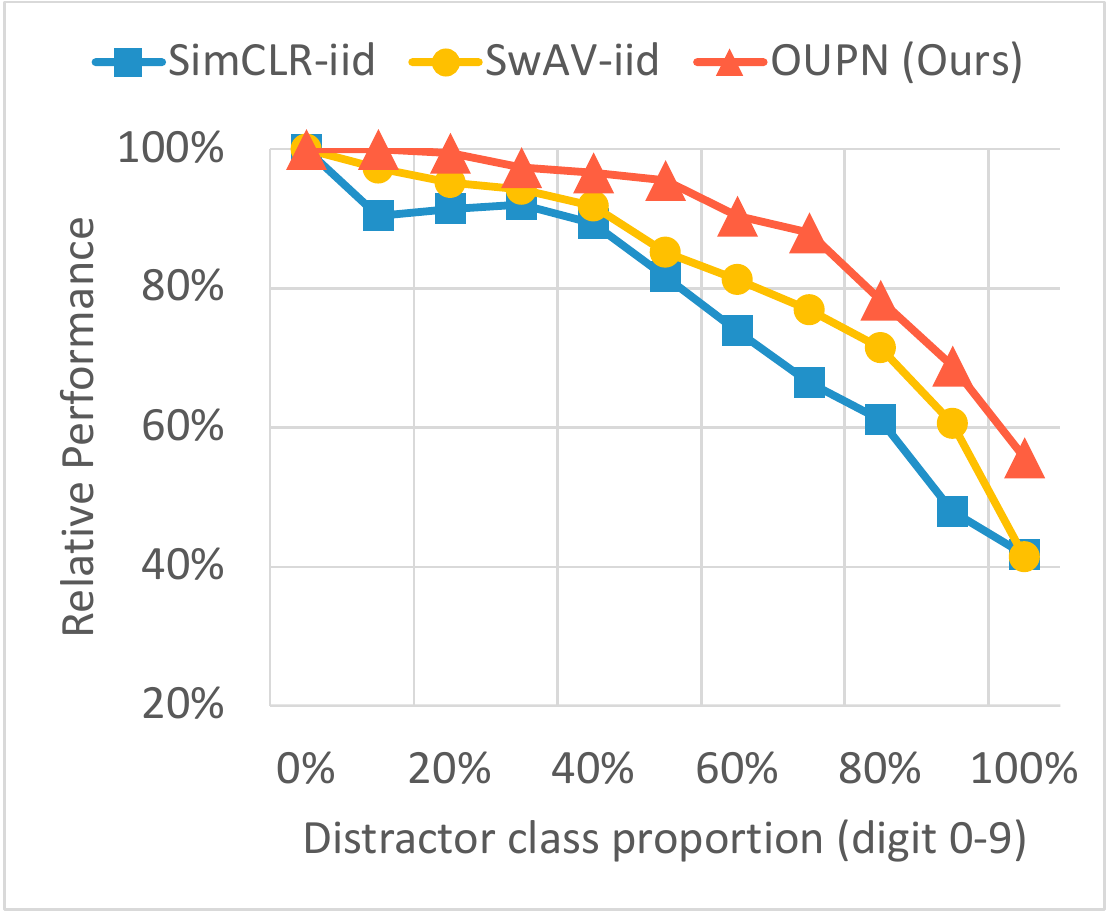}
\quad
\includegraphics[height=3.6cm,trim={0.615cm 0.1cm 0.33cm 0.1cm},clip]{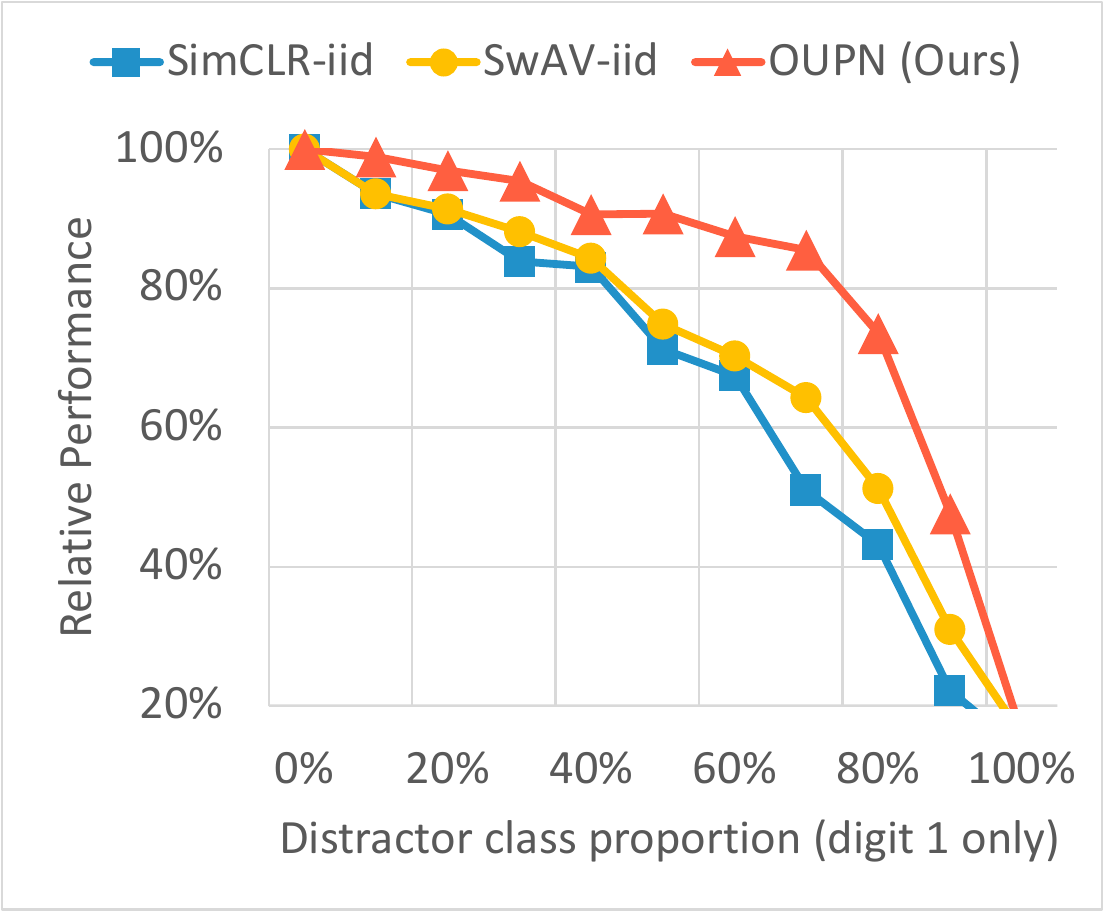}
\fi
\end{minipage}
\ifarxiv
\begin{minipage}[c]{0.35\linewidth}
\caption{Robustness to imbalanced distributions by adding distractors (Omniglot
mixed with MNIST images). Performance is relative to the original and a random
baseline.}
\label{fig:imbalance}
\end{minipage}
\fi
\savespace{-0.25in}
\end{figure}

\savespace{-0.15in}
\paragraph{Results.} The results are reported in Table~\ref{tab:omniglot}. In
both datasets, our method outperforms self-supervised baselines using the same
batch size setting. In \rom{}, our model is able to significantly reduce the
gap between supervised and unsupervised models, however in \rim{} the gap is
still wide, which suggests that our model is still less effective handling more
distinct images of the same semantic class in the online stream.

\savespace{-0.1in}
\paragraph{Effect of imbalanced distribution.}
\looseness=-1000
To achieve a better understanding of why OUPN performs better than other
instance- and clustering-based self-supervised learning methods, here we study
the effect of imbalanced cluster sizes by manipulating the class distribution
in the training episodes. In the first setting, we randomly replace Omniglot
images with MNIST digits, with probability from 0\% to 100\%. For example, at
50\% rate, an MNIST digit is over 300 times more likely to appear compared to
any Omniglot character class, so the episodes are composed of half frequent
classes and half infrequent classes. In the second setting, we randomly replace
Omniglot images with MNIST digit 1 images, which makes the imbalance even
greater. We compared our method to SimCLR and SwAV in the iid setup, since this
is the scenario they were designed for. Results of the two settings are shown
in Fig.~\ref{fig:imbalance}, and our method is shown to be more robust under
imbalanced distribution than SimCLR and SwAV. Compared to clustering-based
methods like SwAV, our prototypes can be dynamically created and updated with
no constraints on the number of elements per cluster. Compared to
instance-based methods like SimCLR, our prototypes sample the contrastive pairs
more equally in terms of representation similarity. We hypothesize that these
model aspects contribute to the differences in robustness.

\ifarxiv
\savespace{-0.1in}
\paragraph{Ablation studies and hyperparameter optimization.}
We study the effect of certain hyperparameters of our model. In
Table~\ref{tab:numproto}, we investigate the effect of the size of the
prototype memory, and whether the model would benefit from a larger memory. It
turns out that as long as the size of the memory is larger than the length of
the input sequence for each gradient update step, it can learn good
representations and the size is not a major determining factor. In
Table~\ref{tab:forget}, we examine whether the memory forgetting parameter is
important to the model. We found that the forgetting rate between 0.99 and
0.995 is the best. 0.999 (closer to no forgetting) results in worse
performance. In Table~\ref{tab:new}, we investigate the effect of various
values for the new cluster loss coefficient. The optimal value is between 0.5
and 1.0. More studies on the effect of $\alpha$, $\tilde{\tau}$,
$\lambda_\textrm{ent}$ and the Beta mean $\mu$ are included in
Appendix~\ref{sec:additionalablation}.
\fi

\ifarxiv
\begin{table}[t]
\savespacefigtop{-0.5in}
\parbox{.33\linewidth}{
\centering
\centering
\caption{Effect of mem. size $K$}
\savespace{0.025in}
\resizebox{!}{1.1cm}{
\begin{small}
\begin{tabular}{l|cc|cc}
\toprule
& \multicolumn{2}{c|}{\rom{}} & \multicolumn{2}{c}{\rro{}} \\
$K$      & AMI       & AP        & AMI       & AP          \\
\midrule
50       & 89.19     & 95.12     & 75.33     & 82.42 \\
100      & \bf 90.54 & 95.83     & 76.70     & 83.51 \\
150      & 90.24     & \bf 95.92 & 77.07     & 84.00 \\
200      & 90.36     & 95.68     & 76.81     & \bf 84.45 \\
250      & 89.87     & 95.69     & \bf 77.83 & 84.33 \\
\bottomrule
\end{tabular}
\end{small}
}
\label{tab:numproto}
}
\parbox{.33\linewidth}{
\centering
\caption{Effect of decay rate $\rho$}
\label{tab:forget}
\resizebox{!}{1.1cm}{
\begin{small}
\begin{tabular}{l|cc|cc}
\toprule
& \multicolumn{2}{c|}{\rom{}} & \multicolumn{2}{c}{\rro{}} \\
$\rho$  & AMI       & AP         & AMI       & AP          \\
\midrule                                                     
0.9     & 51.12  & 64.19         & 65.07     & 75.50       \\
0.95    & 79.78  & 89.30         & 74.33     & 81.92       \\
0.99    & 89.43  & 95.54         & 76.97     & 84.05       \\
0.995   & \bf 90.80 & \bf 95.90  & \bf 77.78 & \bf 85.02   \\
0.999   & 86.27  & 93.69         & 38.89     & 39.37       \\
\bottomrule
\end{tabular}
\end{small}
}
}
\parbox{.33\linewidth}{
\centering
\caption{Effect of $\lambda_\textrm{new}$}
\label{tab:new}
\resizebox{!}{1.1cm}{
\begin{small}
\begin{tabular}{l|cc|cc}
\toprule
& \multicolumn{2}{c|}{\rom{}} & \multicolumn{2}{c}{\rro{}} \\
$\lambda_\textrm{new}$  & AMI       & AP        & AMI       & AP        \\
\midrule
0.0                     & 38.26     & 93.40     & 19.49     & 73.93 \\
0.1                     & 86.60     & 93.50     & 67.25     & 71.69 \\
0.5                     & 89.89     & 95.28     & \bf 78.04 & \bf 84.85 \\
1.0                     & \bf 90.06 & \bf 95.81 & 77.59     & 84.36 \\
2.0                     & 88.74     & 95.73     & 77.62     & 84.72 \\
\bottomrule
\end{tabular}
\end{small}
}
}

\savespace{-0.15in}
\end{table}
\fi

%% file: sections/conclusion.tex
\savespace{-0.1in}
\section{Conclusion}
\savespace{-0.1in}
\label{sec:conclusion}
Our goal is to develop learning procedures for real-world agents who operate
online and in structured, nonstationary environments. Toward this goal, we
develop an online unsupervised algorithm for discovering visual representations
and categories. Unlike standard self-supervised learning, our algorithm embeds
category formation in a probabilistic clustering module that is jointly learned
with the representation encoder. Our clustering is more flexible and supports
learning of new categories with very few examples. At the same time, we
leverage self-supervised learning to acquire semantically meaningful
representations. Our method is evaluated in both synthetic and realistic image
sequences and it outperforms state-of-the-art self-supervised learning
algorithms for both the non-iid sequences we are interested in as well as
sequences transformed to be iid to better match assumptions of the learning
algorithms.

%% file: sections/ack.tex
\section*{Acknowledgments}
Resources used in preparing this research were provided, in part, by the
Province of Ontario, the Government of Canada through CIFAR, and companies
sponsoring the Vector Institute (\url{www.vectorinstitute.ai/\#partners}). This
project is supported by NSERC and the Intelligence Advanced Research Projects
Activity (IARPA) via Department of Interior/Interior Business Center (DoI/IBC)
contract number D16PC00003. The U.S. Government is authorized to reproduce and
distribute reprints for Governmental purposes notwithstanding any copyright
annotation thereon. Disclaimer: The views and conclusions contained herein are
those of the authors and should not be interpreted as necessarily representing
the official policies or endorsements, either expressed or implied, of IARPA,
DoI/IBC, or the U.S. Government.

%% file: sections/appendix.tex

\section{Method Derivation}
\label{sec:derivation}
\subsection{E-step}
\paragraph{Inferring cluster assignments.}
The categorical variable $\hat{y}$ infers the cluster assignment of the current
input example with regard to the existing clusters.
\begin{align}
    \hat{y}_{t,k} &= \Pr(y_t = k | \rvz_{t}, u=0)\\
    &= \frac{\Pr(\rvz_{t} | y_t=k, u=0) \Pr(y_t=k)}{\Pr(\rvz_{t}, u=0)} \\
    &= \frac{w_k f(\rvz_{t}; \rvp_{t,k}, \sigma^2)}{\sum_{k'} w_{k'} f(\rvz_{t}; \rvp_{t,k'}, \sigma^2)}\\
    &= \frac{\exp(\log w_k - d(\rvz_{t}, \rvp_{t,k}) / 2\sigma^2)}{\sum_{k'} \exp(\log w_k' - d(\rvz_{t}, \rvp_{t,k'}) / 2\sigma^2)} \\
    &= \softmax \left(\log w_k - d(\rvz_t, \rvp_{t,k}) / \tau \right),\\
    &= \softmax(v_{t,k}),
\end{align}
where $w_k$ is the mixing coefficient of cluster $k$ and $d(\cdot,\cdot)$ is
the distance function, and $v_{t,k}$ is the logits. In our experiments, $w_k$'s
are kept as constant and $\tau$ is an independent learnable parameter.

\paragraph{Inference on unknown classes.} The binary variable $\hat{u}$
estimates the probability that the current input belongs to a new cluster:
\begin{align}
\hat{u}_t &= \Pr(u_t = 1 | \rvz_{t})\\
      &= \frac{z_0 u_0}{z_0 u_0 + \sum_k w_k f(\rvz_t; \rvp_{t,k}, \sigma^2) (1-u_0)}\\
      &= \frac{1}{1 + \frac{1}{z_0 u_0} \sum_k w_k f(\rvz_t; \rvp_{t,k}, \sigma^2) (1-u_0)}\\
      &= \frac{1}{1 + \exp(\log(\frac{1}{z_0 u_0} \sum_k w_k f(\rvz_t; \rvp_{t,k}, \sigma^2)(1-u_0)))}\\
      &= \frac{1}{1 + \exp(-\log(z_0) - \log(u_0) + \log(1-u_0) + \log(\sum_k w_k f(\rvz_t; \rvp_{t,k}, \sigma^2))}\\
      &= \frac{1}{1 + \exp(-s  + \log(\sum_k \exp(\log(w_k) - d(\rvz_t, \rvp_{t,k}) / 2\sigma^2)))}\\
      &= \sigmoid(s - \log(\sum_k \exp(\log(w_k) - d(\rvz_t, \rvp_{t,k}) / 2\sigma^2)))\\
      &= \sigmoid(s - \log(\sum_k \exp(v_{t,k}))),
\end{align}
where $s = \log(z_0) + \log(u_0) - \log(1-u_0) + m\log(\sigma) + m\log(2\pi)/2$
and $m$ is the input dimension. In our implementation, we use $\max$ here
instead of $\logsumexp$ since we found $\max$ leads to better and more stable
training performance empirically. It can be derived as a lower bound:
\begin{align}
    \hat{u}_t &= \sigmoid(s - \log(\sum_k \exp(\log(w_k) - d(\rvz_t, \rvp_{t,k}) / 2\sigma^2)))\\
    &\ge \sigmoid(s - \log(\max_k \exp(- d(\rvz_t, \rvp_{t,k}) / 2\sigma^2)))\\
    &= \sigmoid(s + \min_k d(\rvz_t, \rvp_{t,k}) / 2\sigma^2)\\
    &= \sigmoid((\min_k d(\rvz_t, \rvp_{t,k}) - \beta) / \gamma),
\end{align}
where $\beta = -2s\sigma^2$, $\gamma=2 \sigma^2$. To make learning more
flexible, we directly make $\beta$ and $\gamma$ as independent learnable
parameters so that we can control the confidence level for predicting unknown
classes.

\subsection{M-step}
\label{sec:appmstep}
Here we infer the posterior distribution of the prototypes $\Pr(\rvp_{t,k} |
\rvz_{1:t})$. We formulate an efficient recursive online update, similar to
Kalman filtering, by incorporating the evidence of the current input $\rvz_t$
and avoiding re-clustering the entire input history. We define
$\hat{\rvp}_{t,k}$ as the estimate of the posterior mean of the $k$-th cluster
at time step $t$, and $\hat{\sigma}^2_{t,k}$ is the estimate of the posterior
variance.

\paragraph{Updating prototypes.}
Suppose that in the E-step we have determined that $y_t=k$. Then the posterior
distribution of the $k$-th cluster after observing $\rvz_t$ is:
\begin{align}
        &~ \Pr(\rvp_{t,k} | \rvz_{1:t}, y_t=k)\\
\propto &~ \Pr( \rvz_{t} | \rvp_{t,k}, y_t=k) \Pr(\rvp_{t,k} | \rvz_{1:t-1})\\
=&~\Pr( \rvz_t | \rvp_{t,k}, y_t=k) \int_{\rvp'} \Pr(\rvp_{t,k} | \rvp_{t-1,k}=\rvp') \Pr(\rvp_{t-1,k}=\rvp' | \rvz_{1:t-1}) \\
\approx &~ f(\rvz_t; \rvp_{t,k}, \sigma^2) \int_{\rvp'} f(\rvp_{t,k}; \rvp', \sigma_{t,d}^2) f(\rvp'; \hat{\rvp}_{t-1,k}, \hat{\sigma}_{t-1,k}^2) \\
=       &~ f(\rvz_t; \rvp_{t,k}, \sigma^2) f(\rvp_{t,k}; \hat{\rvp}_{t-1,k}, \sigma_{t,d}^2 + \hat{\sigma}_{t-1,k}^2).
\end{align}
If we assume that the transition probability distribution $\Pr(\rvp_{t,k} |
\rvp_{t-1,k})$ is a zero-mean Gaussian with variance $\sigma_{t,d}^2 = (1/\rho
- 1) \hat{\sigma}_{t-1,k}^2$, where $\rho
\in (0, 1]$ is some constant that we defined to be the memory decay
coefficient, then the posterior estimates are:
\begin{align}
    \hat{\rvp}_{t,k}|_{y_t=k} = \frac{\rvz_t \hat{\sigma}_{t-1,k}^2 / \rho + \hat{\rvp}_{t-1,k} \sigma^2}{\sigma^2 + \hat{\sigma}_{t-1,k}^2 / \rho}, \quad
    \hat{\sigma}_{t,k}^2|_{y_t=k} = \frac{\sigma^2\hat{\sigma}_{t-1,k}^2 / \rho}{\sigma^2+ \hat{\sigma}_{t-1,k}^2 / \rho}.
\end{align}
If $\sigma^2=1$, and $\hat{c}_{t,k} \equiv 1/\hat{\sigma}_{t,k}^2$,
$\hat{c}_{t-1,k} \equiv 1/\hat{\sigma}_{t-1,k}^2$, it turns out we can
formulate the update equation as follows, and $\hat{c}_{t,k}$ can be viewed as
a count variable for the number of elements in each estimated cluster, subject
to the decay factor $\rho$ over time:
\begin{align}
    \hat{c}_{t,k}|_{y_t=k} &= \rho \hat{c}_{t-1,k} + 1, \\
    \hat{\rvp}_{t,k}|_{y_t=k} &= \rvz_t \frac{1}{\hat{c}_{t,k}|_{y_t=k}} + \hat{\rvp}_{t-1,k}\frac{\rho \hat{c}_{t-1,k}}{\hat{c}_{t,k}|_{y_t=k}}.
\end{align}
If $y_t \neq k$, then the prototype posterior distribution simply gets diffused
at timestep $t$:
\begin{align}
    \Pr(\rvp_{t,k} | z_{1:t}, y_t \neq k) &\approx f(\rvp_{t,k}; \hat{\rvp}_{t-1,k}, \hat{\sigma}_{t-1,k}^2 / \rho)\\
    \hat{c}_{t,k}|_{y_t \neq k} &= \rho \hat{c}_{t-1,k}, \\
    \hat{\rvp}_{t,k}|_{y_t \neq k} &= \hat{\rvp}_{t-1,k}.
\end{align}
Finally, our posterior estimates at time $t$ are computed by taking the
expectation over $y_t$:
\begin{align}
\hat{c}_{t,k} &= \mathop{\mathbb{E}}_{y_t} [\hat{c}_{t,k}|_{y_t}] \\
&= \hat{c}_{t,k}|_{y_t=k} \Pr(y_t=k | \rvz_t) + \hat{c}_{t,k}|_{y_t \neq k} \Pr(y_t \neq k | \rvz_t) \\
&= (\rho \hat{c}_{t-1,k} + 1) \hat{y}_{t,k} (1-\hat{u}_{t,k}) + \rho \hat{c}_{t-1,k}(1- \hat{y}_{t,k} (1-\hat{u}_{t,k})), \\
&= \rho \hat{c}_{t-1,k} + \hat{y}_{t,k} (1-\hat{u}_{t,k}), \\
\hat{\rvp}_{t,k} &= \mathop{\mathbb{E}}_{y_t} [\hat{\rvp}_{t,k} |_{y_t}] \\ 
&= \hat{\rvp}_{t,k} |_{y_t=k} \Pr(y_t=k | \rvz_t) + \hat{\rvp}_{t,k} |_{y_t \neq k} \Pr(y_t \neq k | \rvz_t) \\
&= \left(\rvz_t \frac{1}{\hat{c}_{t,k}|_{y_t=k}} + \hat{\rvp}_{t-1,k}\frac{\rho \hat{c}_{t-1,k}}{\hat{c}_{t,k}|_{y_t=k}} \right) \hat{y}_{t,k}(1-\hat{u}_{t,k}) + \hat{\rvp}_{t-1,k}(1-\hat{y}_{t,k}(1-\hat{u}_{t,k}))\\
&= \rvz_t \frac{\hat{y}_{t,k} (1-\hat{u}_{t,k})}{\rho\hat{c}_{t-1,k} + 1} + \hat{\rvp}_{t-1,k} \left( 1- \hat{y}_{t,k}(1-\hat{u}_{t,k}) + \hat{y}_{t,k}(1-\hat{u}_{t,k}) \frac{\rho \hat{c}_{t-1,k}}{\rho\hat{c}_{t-1,k} + 1} \right)\\
&= \rvz_t \frac{\hat{y}_{t,k} (1-\hat{u}_{t,k})}{\rho\hat{c}_{t-1,k} + 1} + \hat{\rvp}_{t-1,k} \left( 1-  \frac{\hat{y}_{t,k}(1-\hat{u}_{t,k})}{\rho\hat{c}_{t-1,k} + 1} \right).
\end{align}
Since $\hat{c}_{t,k}$ is also our estimate on the number of elements in each
cluster, we can use it to estimate the mixture weights,
\begin{align}
    \hat{w}_{t,k} = \frac{\hat{c}_{t,k}}{\sum_{k'} \hat{c}_{t,k}}.
\end{align}
Note that in our experiments the mixture weights are not used and we assume
that each cluster has an equal mixture probability.

\section{Experiment Details}
\label{sec:impl}
We provide additional implementation details in Tab.~\ref{tab:rooms_params},
\ref{tab:saycam_params}, \ref{tab:omniglot_params} and
\ref{tab:imagenet_params}.

\paragraph{\rro{}.} For baseline self-supervised learning methods, learning rate
is scaled based on batch size $/ 256 \times 0.3$ using the default LARS
optimizer with cosine learning rate decay and 1 epoch of linear learning rate
warmup. We trained for a total of 10,240,000 examples. So the total number of
training steps is 10,240,000 $/$ batch size. For our proposed model, we used
the batch size of 50 and trained for a total of 80,000 steps (4,000,000
examples), using the Adam optimizer and staircase learning rate decay starting
from $10^{-3}$, with 10$\times$ learning rate decay at 40k and 60k training
steps. All data augmentation parameters are the same as the original SimCLR
paper, except that in RoamingRooms the minimum crop area is changed to 0.2
instead of the default 0.08. Other details can be found in
Table~\ref{tab:rooms_params}.

\begin{table}[t]
\centering
\caption{Experiment details for \rro{}}
\begin{small}
\begin{tabular}{l|l}
\toprule
Hyperparameter                  & Values               \\
\midrule
$\tau$ init                     & 0.1                  \\
$\beta$ init                    & -12.0                \\
$\gamma$ init                   & 1.0                  \\
Num prototypes $K$              & 150                  \\
Memory decay $\rho$             & 0.995                \\
Beta mode $\mu$                 & 0.5                  \\
Entropy loss $\lambda_\textrm{ent}$ & 0.0              \\
New cluster loss $\lambda_\textrm{new}$ & 0.5          \\
Threshold $\alpha$              & 0.5                  \\
Pseudo label temperature ratio $\tilde{\tau} / \tau$ & 0.1   \\
\bottomrule
\end{tabular}
\end{small}
\label{tab:rooms_params}
\end{table}

\paragraph{SAYCam.}
Data augmentation is slightly different from the standard static image setting.
We found that there were a lot of blurred and shaking frames in the videos.
Therefore, we added random rotation, motion blur and Gaussian blur in the data
augmentation procedure for all methods (including the baselines). Motion blur
is generated with a uniformly random direction between [0$\degree$,
360$\degree$), with the length to be 5\% of the image height, and Gaussian blur
is generated by a blur kernel of 5\% of the image height with the standard
deviation to be uniform between [0.1, 1.2). Same to all the baselines, our
SAYCam model also applies two different augmentations on each image in the
input pair.

For baseline self-supervised learning methods, the learning rate is scaled
based on the batch size $/ 256 \times 0.3$, or 0.0879 (batch size = 75), using
the default LARS optimizer with cosine learning rate decay and 1 epoch of
linear learning rate warmup. We trained the models for a total of 16 epochs.
The total number of training steps is 31,568 (1,973 steps per epoch). For the
TC-S model, we used the Adam optimizer with learning rate 1e-3 and batch size
75. We trained it for 38k steps with 10$\times$ learning rate decays at 25k and
35k. For our model, we used the Adam optimizer with learning rate 1e-3, for a
total of 30k training steps, with a 10$\times$ learning rate decay at 20k
steps.

For $\hat{y}$ and $\hat{u}$, we found it was helpful to sample binary values
for the two variables in the forward pass, and use gradient straight-through
estimator in the backward pass. This modification was only applied on SAYCam
experiments. Other details can be found in Table~\ref{tab:saycam_params}.

\paragraph{\rom{} and \rim{}.}
For baseline self-supervised learning methods on \rom{}, the learning rate is
scaled based on the batch size $/ 256 \times 0.5$, or 0.293 (batch size = 150),
using the default LARS optimizer with cosine learning rate decay and 10 epochs
of linear learning rate warmup. We trained the model for a total of 1,000
epochs. The total number of training steps is 527,000 (527 per epoch).

For baseline self-supervised learning methods on \rim{}, the learning rate is
scaled based on the batch size $/ 256 \times 0.3$, or 0.05625 (batch size =
48), using the default LARS optimizer with cosine learning rate decay and 1
epoch of linear learning rate warmup. We trained the models for a total of 10
epochs. The total number of training steps is 93,480 (9,348 per epoch).

For our model on both datasets, we train using the Adam optimizer with learning
rate 1e-3, for a total of 80k training steps, with $10\times$ learning rate
decay at 40k and 60k. More implementation details can be found in
Table~\ref{tab:omniglot_params} and \ref{tab:imagenet_params}.

\begin{table}[t]
\centering
\caption{Experiment details for SAYCam}
\begin{small}
\begin{tabular}{l|l}
\toprule
Hyperparameter                  & Values               \\
\midrule
Random motion blur              & 30\% probability \\
Random Gaussian blur              & 20\% probability \\
Random rotation                 & uniform between -15$\degree$ and 15$\degree$ \\
$\tau$ init                     & 0.1                  \\
$\beta$ init                    & -12.0                \\
$\gamma$ init                   & 1.0                  \\
Num prototypes $K$              & 75                  \\
Memory decay $\rho$             & 0.99                 \\
Beta mean $\mu$                 & 0.6 (mode=0.7)    \\
Entropy loss $\lambda_\textrm{ent}$ & 0.0              \\
New cluster loss $\lambda_\textrm{new}$ & 0.3          \\
Threshold $\alpha$              & 0.5                  \\
Pseudo label temperature ratio $\tilde{\tau} / \tau$ & 0.0 (i.e. one-hot pseudo labels)     \\
\bottomrule
\end{tabular}
\end{small}
\label{tab:saycam_params}
\end{table}

\begin{table}[t]
\centering
\caption{Experiment details for \rom{}}
\begin{small}
\begin{tabular}{l|l}
\toprule
Hyperparameter                  & Values               \\
\midrule
$\tau$ init                     & 0.1                  \\
$\beta$ init                    & -12.0                \\
$\gamma$ init                   & 1.0                \\
Num prototypes $K$              & 150                  \\
Memory decay $\rho$             & 0.995                \\
Beta mean $\mu$                 & 0.5                  \\
Entropy loss $\lambda_\textrm{ent}$ & 1.0              \\
New cluster loss $\lambda_\textrm{new}$ & 1.0          \\
Threshold $\alpha$              & 0.5                  \\
Pseudo label temperature ratio $\tilde{\tau} / \tau$ & 0.2 \\
\bottomrule
\end{tabular}
\end{small}
\label{tab:omniglot_params}
\end{table}

\begin{table}[t]
\centering
\caption{Experiment details for \rim{}}
\begin{small}
\begin{tabular}{l|l}
\toprule
Hyperparameter                  & Values               \\
\midrule
$\tau$ init                     & 0.1                  \\
$\beta$ init                    & -12.0                \\
$\gamma$ init                   & 1.0                  \\
Num prototypes $K$              & 600                  \\
Memory decay $\rho$             & 0.99                 \\
Beta mean $\mu$                 & 0.5                  \\
Entropy loss $\lambda_\textrm{ent}$ & 0.5              \\
New cluster loss $\lambda_\textrm{new}$ & 0.5          \\
Threshold $\alpha$              & 0.5                  \\
Pseudo label temperature ratio $\tilde{\tau} / \tau$ & 0.0 (i.e. one-hot pseudo labels)     \\
\bottomrule
\end{tabular}
\end{small}
\label{tab:imagenet_params}
\end{table}

\begin{table}[t]
\begin{center}
\caption{Unsupervised iid learning on Omniglot using an MLP}
\label{tab:iidmlp}
\begin{small}
\begin{tabular}{cccc}
\toprule
Method               & 3-NN Error         & 5-NN Error         & 10-NN Error        \\
\midrule                                                                              
VAE~\citep{dirvae}    & 92.34$\pm$0.25     & 91.21$\pm$0.18     & 88.79$\pm$0.35     \\
SBVAE~\citep{dirvae}  & 86.90$\pm$0.82     & 85.10$\pm$0.89     & 82.96$\pm$0.64     \\
DirVAE~\citep{dirvae} & 76.55$\pm$0.23     & 73.81$\pm$0.29     & 70.95$\pm$0.29     \\
CURL~\citep{curl}     & 78.18$\pm$0.47     & 75.41$\pm$0.34     & 72.51$\pm$0.46     \\
SimCLR~\citep{simclr} & 44.35$\pm$0.55     & 42.99$\pm$0.55     & 44.93$\pm$0.55     \\
SwAV~\citep{swav}     & \bf 42.66$\pm$0.55 & \bf 42.08$\pm$0.55 & 44.78$\pm$0.55     \\
OUPN (Ours)          & 43.75$\pm$0.55     & \bf 42.13$\pm$0.55 & \bf 43.88$\pm$0.55 \\
\bottomrule
\end{tabular}
\end{small}
\end{center}
\end{table}

\subsection{Metric Details}
For each method, we used the same nearest centroid algorithm for online
clustering. For unsupervised readout, at each timestep, if the closest centroid
is within threshold $\alpha$, then we assign the new example to the cluster,
otherwise we create a new cluster. For supervised readout, we assign examples
based on the class label, and we create a new cluster if and only if the label
is a new class. Both readout procedures will provide us a sequence of class
IDs, and we will use the following metrics to compare our predicted class IDs
and groundtruth class IDs. Both metrics are designed to be threshold invariant.

\paragraph{AMI.}
For unsupervised evaluation, we consider the adjusted mutual information score.
Suppose we have two clustering $U = \{U_i\}$ and $V = \{V_j\}$, and $U_i$ and
$V_j$ are set of example IDs, and $N$ is the total number of examples. $U$ and
$V$ can be viewed as discrete probability distribution over cluster IDs.
Therefore, the mutual information score between $U$ and $V$ is:
\begin{align}
    \textrm{MI}(U, V) &= \sum_{i=1}^{\lvert U \rvert} \sum_{j=1}^{\lvert V \rvert} \frac{ \lvert U_i
    \cap V_j\rvert }{N} \log \left( \frac{N \lvert U_i \cap V_j\rvert }{\lvert U_i \rvert \lvert V_j
    \rvert} \right) \\ &= \sum_{i=1}^R \sum_{j=1}^C \frac{n_{ij}}{N} \log \left( \frac{N n_{ij}}{a_i
    b_j} \right).
\end{align}
The adjusted MI
score\footnote{\url{https://scikit-learn.org/stable/modules/generated/sklearn.metrics.adjusted_mutual_info_score.html}}
normalizes the range between 0 and 1, and subtracts the baseline from random
chance:
\begin{align}
    \textrm{AMI}(U, V) &= \frac{MI(U, V) - \mathbb{E}[MI(U,V)]}{\frac{1}{2}(H(U) + H(V))  -
    \mathbb{E}[MI(U,V)]},
\end{align}
where $H(\cdot)$ denotes the entropy function, and $\mathbb{E}[MI(U,V)]$ is the
expected mutual information by chance
\footnote{\url{https://en.wikipedia.org/wiki/Adjusted_mutual_information}}.
Finally, for each model, we sweep the threshold $\alpha$ to get a threshold invariant score:
\begin{align}
\textrm{AMI}_{\max} = \max_\alpha \textrm{AMI}(y, \hat{y}(\alpha)).
\end{align}

\paragraph{AP.}
For supervised evaluation, we used the AP metric. The AP metric is also
threshold invariant, and it takes both output $\hat{u}$ and $\hat{y}$ into
account. First it sorts all the prediction based on its unknown score $\hat{u}$
in ascending order. Then it checks whether $\hat{y}$ makes the correct
prediction. For the N top ranked instances in the sorted list, it computes:
precision@N and recall@N among the known instances.
\begin{itemize}
    \item precision@$N$ = $\frac{1}{N} \sum_n \mathbbm{1}[\hat{y}_n = y_n]$, 
    \item recall@$N$ = $\frac{1}{K} \sum_n \mathbbm{1}[\hat{y}_n = y_n]$,
\end{itemize}
where $K$ is the true number of known instances among the top N instances.
Finally, AP is computed as the area under the curve of (y=precision@N,
x=recall@N). For more details, see Appendix A.3 of \citet{www}.

\section{Additional Experimental Results}

\subsection{Comparison to Reconstruction-Based Methods}
We additionally provide Tab.~\ref{tab:iidmlp} to show a comparison with
CURL~\citep{curl} in the iid setting. We used the same MLP architecture and
applied it on the Omniglot dataset using the same data split.
Reconstruction-based methods lag far behind self-supervised learning methods.
Our method is on par with SimCLR and SwAV.

\begin{table}[t]
\begin{minipage}{.49\linewidth}
\centering
\caption{Effect of threshold $\alpha$}
\label{tab:threshold}
\resizebox{!}{1.5cm}{
\begin{small}
\begin{tabular}{l|cc|cc}
\toprule
& \multicolumn{2}{c|}{\rom{}} & \multicolumn{2}{c}{\rro{}} \\
$\alpha$  & AMI       & AP        & AMI       & AP         \\
\midrule
0.3       & 82.75     & 90.57     & 52.60     & 58.71      \\
0.4       & 81.59     & 90.94     & 59.69     & 66.11      \\
0.5       & \bf 89.65 & \bf 95.22 & \bf 77.96 & \bf 84.34  \\
0.6       & 87.01     & 93.87     & 64.65     & 69.49      \\
0.7       & 86.08     & 92.94     & 66.60     & 73.54      \\
\bottomrule
\end{tabular}
\end{small}
}
\end{minipage}
\begin{minipage}{.49\linewidth}
\centering
\caption{Effect of $\tilde{\tau}$}
\label{tab:temperature}
\resizebox{!}{1.5cm}{
\begin{small}
\begin{tabular}{l|cc|cc}
\toprule
& \multicolumn{2}{c|}{\rom{}} & \multicolumn{2}{c}{\rro{}}       \\
$\tilde{\tau}$ / $\tau$  & AMI        & AP        & AMI       & AP        \\
\midrule
0.05            & 89.23      & 95.01     & 77.44     & 84.38 \\
0.10            & 89.71      & 95.21     & \bf 77.89 & \bf 84.99 \\
0.20            & \bf 89.78  & \bf 95.31 & 77.82     & 84.57 \\
0.50            & 89.40      & 95.13     & 76.81     & 83.90 \\
1.00            & 89.62      & 95.16     & 0.00      & 19.91 \\
\bottomrule
\end{tabular}
\end{small}
}
\end{minipage}
\begin{minipage}{.49\linewidth}
\vspace{0.2in}
\centering
\caption{Effect of $\lambda_\textrm{ent}$}
\label{tab:entropy}
\resizebox{!}{1.5cm}{
\begin{small}
\begin{tabular}{l|cc|cc}
\toprule
& \multicolumn{2}{c|}{\rom{}} & \multicolumn{2}{c}{\rro{}}               \\
$\lambda_\textrm{ent}$  & AMI       & AP        & AMI    & AP            \\
\midrule
0.00                    & 82.45     & 90.66     & \bf 76.64  & \bf 84.11 \\
0.25                    & 87.31     & 93.85     & 76.61  & 83.16 \\
0.50                    & 87.98     & 94.21     & 75.46  & 81.78 \\
0.75                    & 88.77     & 94.74     & 74.76  & 79.91 \\
1.00                    & \bf 89.70 & \bf 95.14 & 75.32  & 80.29 \\
\bottomrule
\end{tabular}
\end{small}
}
\end{minipage}
\begin{minipage}{.49\linewidth}
\vspace{0.2in}
\centering
\caption{Effect of mean $\mu$ of the Beta prior}
\label{tab:target}
\resizebox{!}{1.5cm}{
\begin{small}
\begin{tabular}{l|cc|cc}
\toprule
& \multicolumn{2}{c|}{\rom{}} & \multicolumn{2}{c}{\rro{}}               \\
$\mu$  & AMI       & AP        & AMI       & AP            \\
\midrule
0.3    & 84.14     & 93.19     & 68.75     & 72.58  \\
0.4    & 86.59     & 93.10     & 69.19     & 73.86  \\
0.5    & \bf 89.89 & \bf 95.24 & \bf 77.61 & \bf 84.64  \\
0.6    & 85.93     & 93.81     & 64.21     & 73.23  \\
0.7    & 26.22     & 92.08     & 48.58     & 64.28  \\
\bottomrule
\end{tabular}
\end{small}
}
\end{minipage}
\end{table}

\subsection{Additional Studies on Hyperparameters}
\label{sec:additionalablation}

In Table~\ref{tab:threshold}, the threshold parameter is found to be the best
at 0.5. However, this could be correlated with how frequently the frames are
sampled.

In Table~\ref{tab:temperature}, we found that the soft distillation loss is
beneficial and slightly improves the performance compared to hard distillation.

In Table~\ref{tab:entropy}, the entropy loss we introduced leads to a
significant improvement on the Omniglot dataset but not on the RoamingRooms
dataset.

The Beta $\mu$ is computed as the following: Suppose $a$ and $b$ are the
parameters of the Beta distribution, and $\mu$ is the mean. We fix $a = 4 \mu$
and $b= 4 - a$. In Table~\ref{tab:target}, we found that the
mean of the Beta prior is the best at 0.5. It has more impact on the \rro{}
dataset, and has less impact on the \rom{} dataset. This parameter could be
influenced by the total number of clusters in each sequence.

\section{Additional Visualization Results}
\label{sec:additional_viz}

\begin{figure}[t]
\centering
\fbox{
\includegraphics[width=\textwidth]{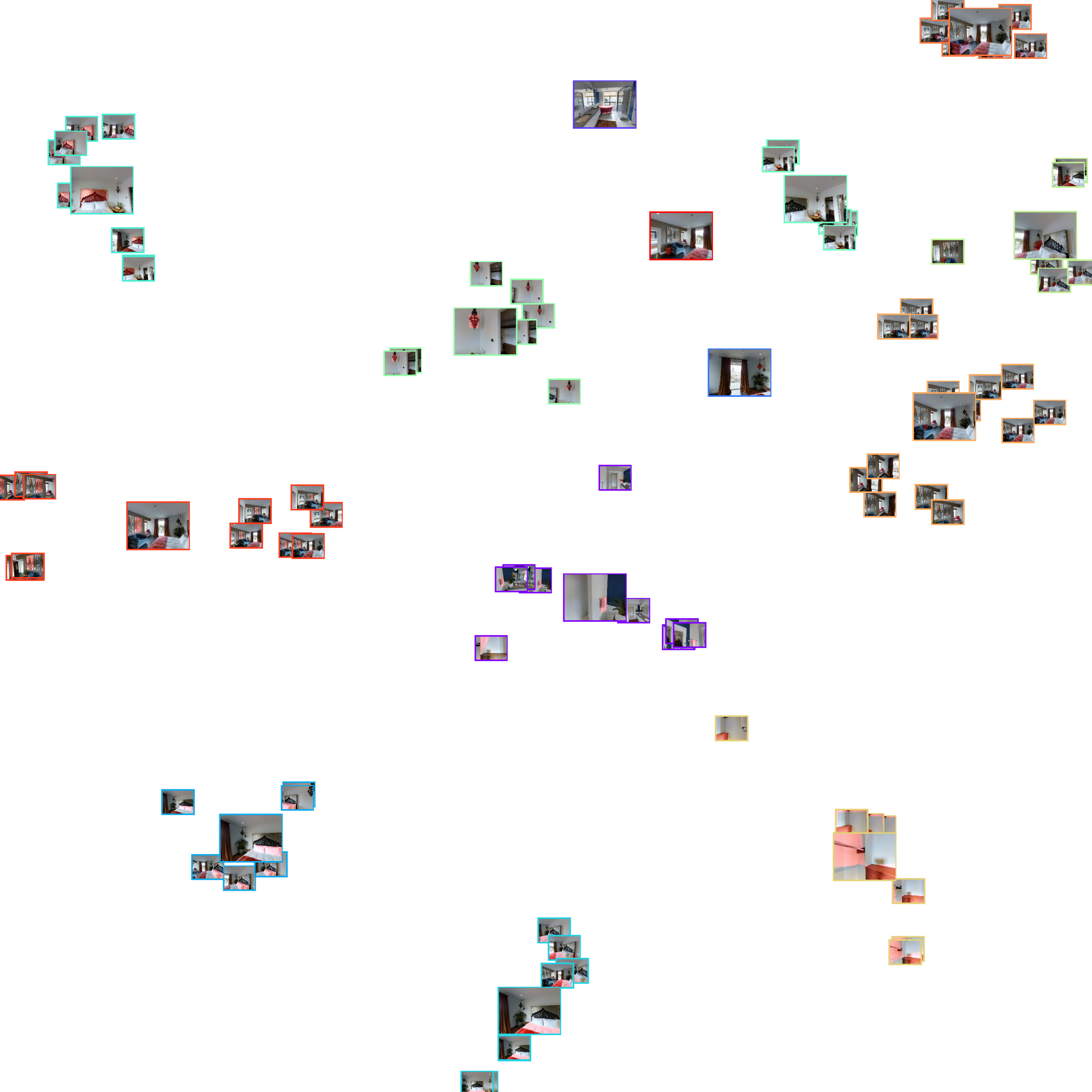}
}
\caption{Embeddings and clustering outputs of an example episode (1).
Embeddings are extracted from the trained CNN and projected to 2D space using
t-SNE~\citep{tsne}. The main object in each image is highlighted in a red mask.
The nearest example to each cluster centroid is enlarged. Image border colors
indicate the cluster assignment.}
\label{fig:clusterembed1}
\end{figure}

\begin{figure}
\centering
\fbox{
\includegraphics[width=\textwidth]{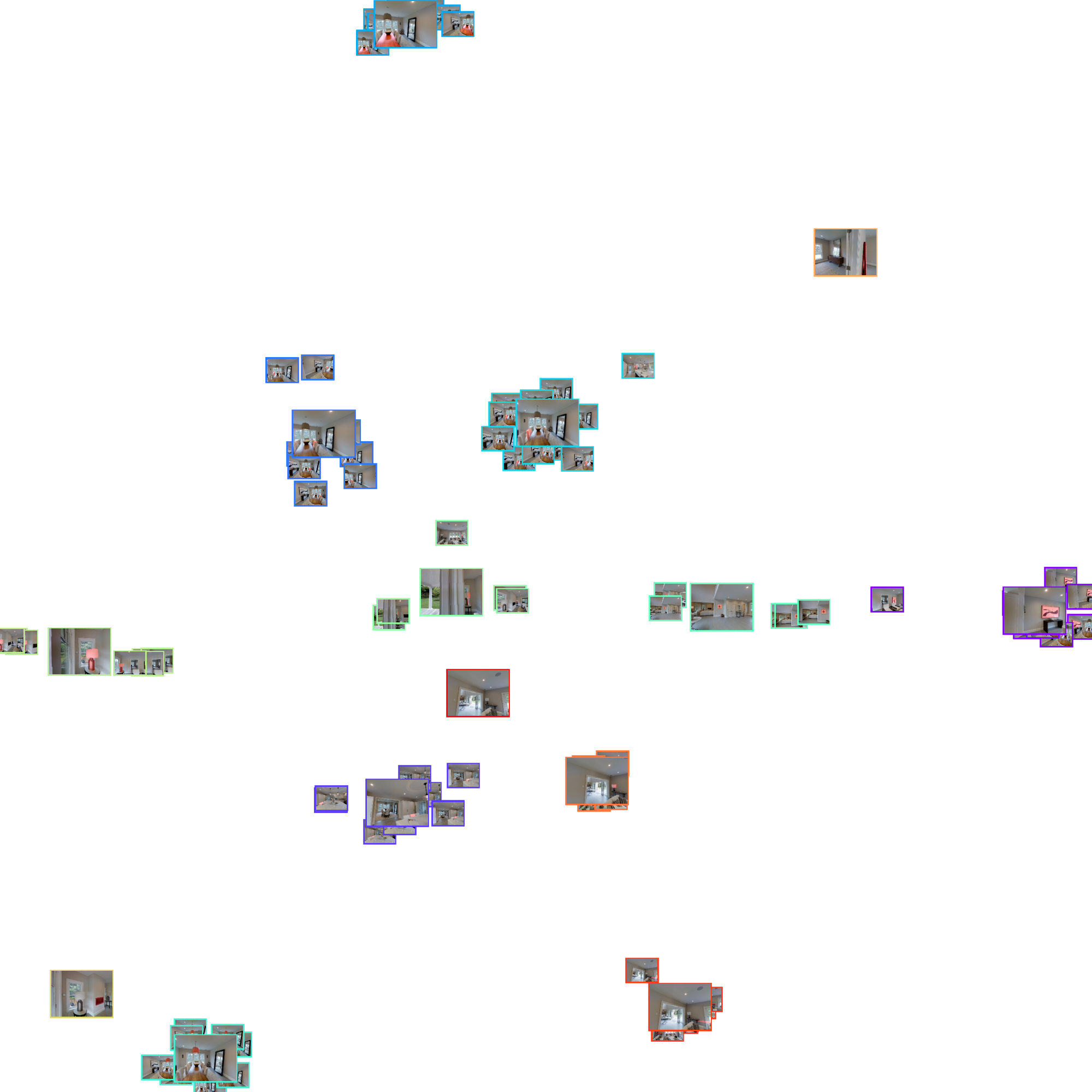}
}
\caption{Embeddings and clustering outputs of another example episode (2).}
\label{fig:clusterembed2}
\end{figure}

We visualize the clustering mechanism and the learned image embeddings on
\rro{} in Fig.~\ref{fig:clusterembed1} and \ref{fig:clusterembed2}. The results
suggest that our model can handle a certain level of view point changes by
grouping different view points of the same object into a single cluster. It
also shows that our model is instance-sensitive: for example, the headboard,
pillows, and the blanket are successfully separated.


\begin{figure}[t]
\centering
\fbox{
\includegraphics[width=\textwidth,trim={3cm 3cm 3cm
3cm},clip]{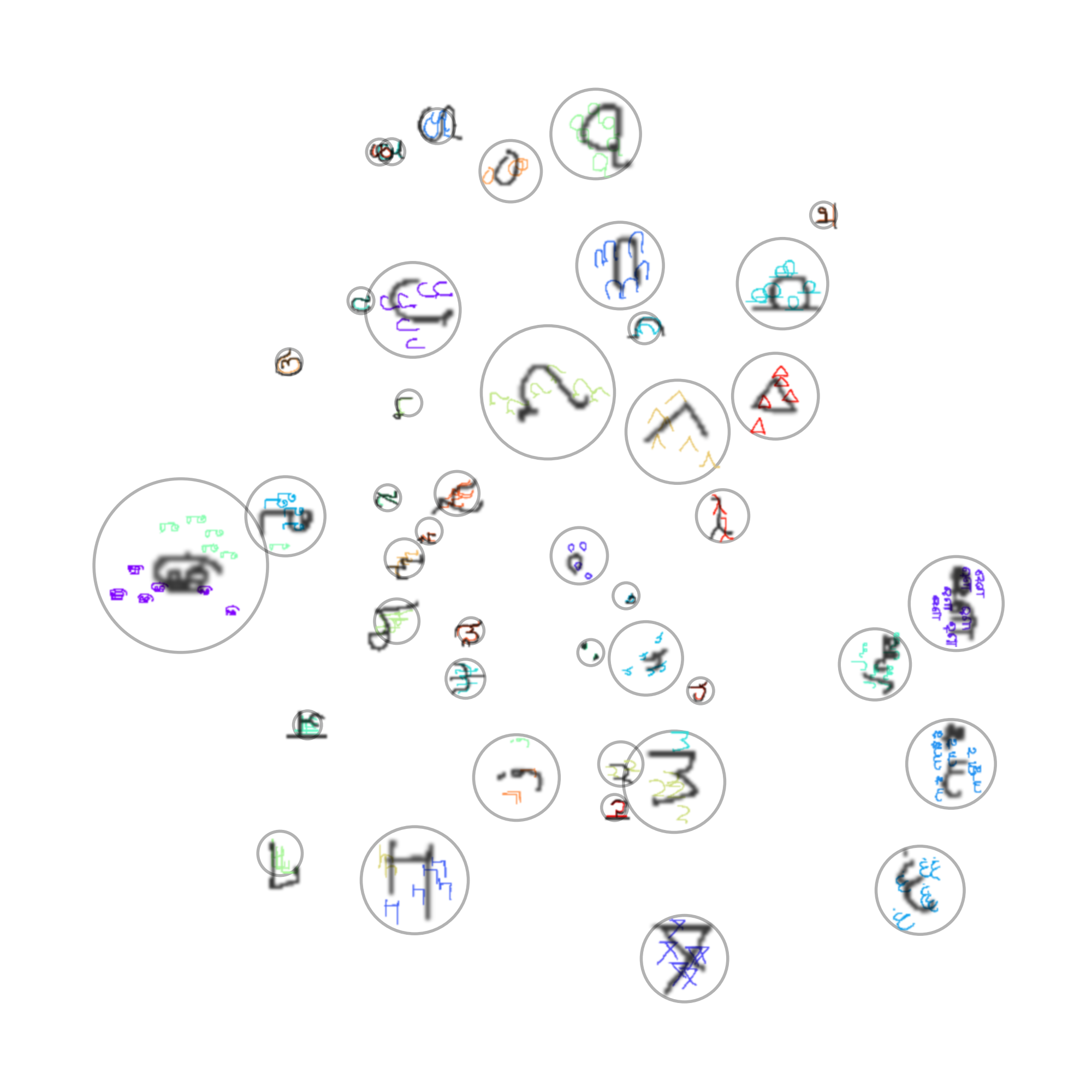} }
\caption{Embedding visualization of an unsupervised training episode of \rom{}.
Different colors denote the ground-truth class IDs.}
\label{fig:clusterembed3}
\end{figure}

\begin{figure}[t]
\centering
\fbox{
\includegraphics[width=\textwidth,trim={3cm 3cm 3cm
3cm},clip]{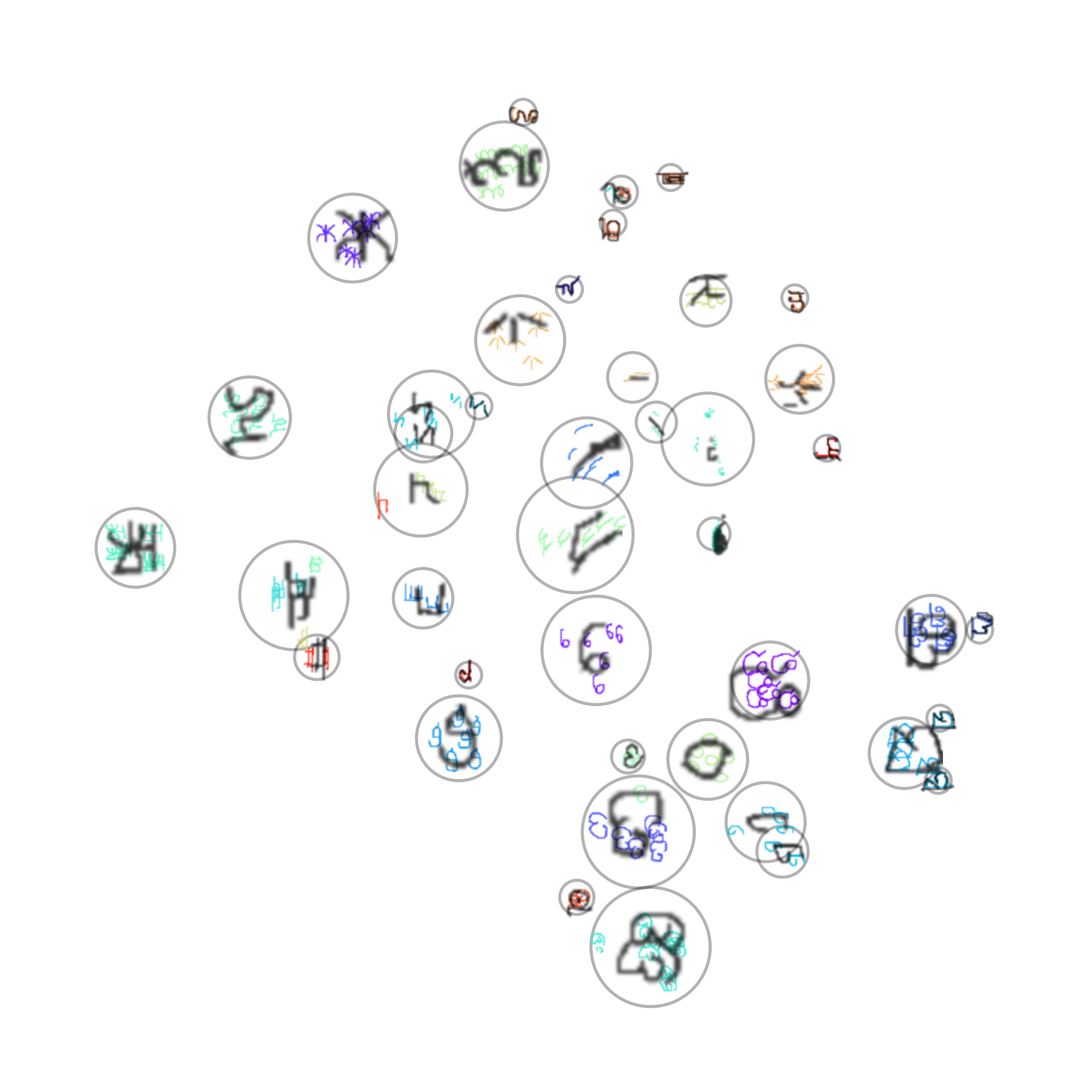} }
\caption{Embedding visualization of an test episode of \rom{}.}
\label{fig:clusterembed4}
\end{figure}

In Fig.~\ref{fig:clusterembed3} and \ref{fig:clusterembed4}, we visualize the
learned categories in
\rom{} using t-SNE~\citep{tsne}. Different colors represent different
ground-truth classes. Our method is able to learn meaningful embeddings and
roughly group items of similar semantic meanings together.

